\documentclass[journal]{IEEEtran}
\usepackage{amsmath,amsfonts}
\usepackage{algorithmic}
\usepackage{algorithm}
\usepackage{array}
\usepackage{amsmath}
\usepackage{array}
\usepackage{bm}
\usepackage{booktabs}
\usepackage{textcomp}
\usepackage{stfloats}
\usepackage{multirow}
\usepackage{url}
\usepackage{verbatim}
\usepackage{cite}
\usepackage{graphicx}
\usepackage{subfigure}
\usepackage{threeparttable}
\usepackage{tabularx}
\usepackage{lineno}
\usepackage{enumitem}
\usepackage{soul,color,xcolor}
\usepackage{hyperref}
\usepackage{float}
\begin{document}

\title{Reliability-Adaptive Multimodal Relative Pose Tracking for Underwater Leader--Follower Robots}


\author{Mingyang Yang, Zeyu Sha, Xiaorui Wang, Jun Wang, David Chen Wang and Feitian Zhang$^*$
\thanks{The authors are with the Department of Robotics and Manufacturing Engineering, College of Advanced Manufacturing and Robotics, and the State Key Laboratory of Turbulence and Complex Systems, Peking University, Beijing, 100871, China (emails:
        {mingyangyang@stu.pku.edu.cn},  {schahzy@stu.pku.edu.cn}, {jnswxr@stu.pku.edu.cn}, {wjun@pku.org.cn}, {6david035@gmail.com} and {feitian@pku.edu.cn}).}%
\thanks{* Send all correspondence to Feitian Zhang.}
}

\maketitle

\begin{abstract} 
Underwater leader-follower formation control requires continuous relative-pose estimation despite degraded sensing conditions and limited onboard computation. This paper presents a reliability-adaptive multimodal perception framework for six-degree-of-freedom relative-pose tracking between two moving underwater robots. The framework integrates complementary visual, acoustic-ranging, and hydrodynamic measurements while accounting for their distinct physical characteristics and time-varying reliability. It adaptively weights modality contributions and models temporal dependencies with linear computational complexity, enabling efficient pose estimation on embedded hardware. Extensive experiments were conducted using a custom-built AUV platform, with 648 pool trials spanning two leader vehicle types, three illumination conditions, and nine trajectory patterns. Evaluation yields an average translational error of 3.8~cm and an average orientation error of 2.4$^\circ$. On an onboard NVIDIA Jetson AGX Orin, the fusion network achieves an inference latency of 17.62~ms, providing a 2.69-fold speedup over the evaluated Transformer-based fusion baseline. Closed-loop experiments further demonstrate stable formation maintenance with a moving leader under onboard computational constraints. 

A multimodal dataset containing measurements from all 648 trials will be made publicly available.
\end{abstract}

\begin{IEEEkeywords}
Multimodal sensor fusion, relative pose tracking, leader-follower formation, autonomous underwater vehicles, adaptive sensor weighting, embedded perception.
\end{IEEEkeywords}

\section{Introduction}
\subsection{Background and Motivation}
\begin{figure}[thpb]
 \centering
 \includegraphics[width=0.485\textwidth]{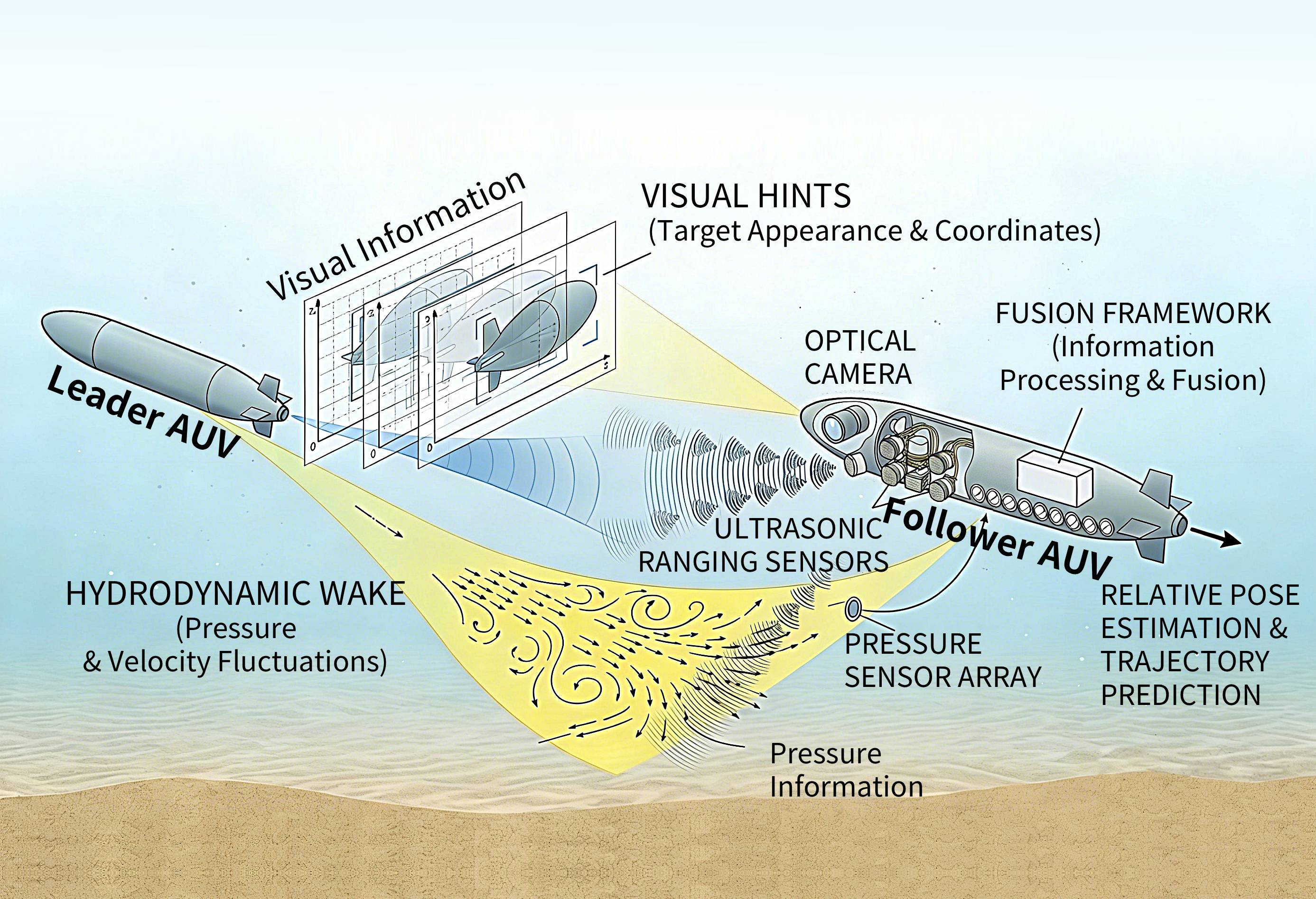}
 \caption{Schematic of the proposed multimodal fusion framework for underwater leader-follower formation control. The follower AUV obtains complementary information about the leader through visual observations, active ultrasonic ranging measurements, and hydrodynamic wake measurements. These measurements are acquired using a dedicated sensor suite comprising a monocular camera, ultrasonic sensors, and a pressure sensor array, respectively. The fusion framework processes these inputs to estimate the leader's relative pose and support real-time formation control.}
\label{IntroFig}
\end{figure}

\IEEEPARstart{A}{utonomous} underwater vehicles (AUVs) play important roles in ocean exploration, including bathymetric surveying \cite{RALBathSurvey}, infrastructure inspection \cite{JOEStructureIns}, environmental monitoring \cite{TROENVMON}, and polar studies \cite{polarstudy}. However, single-vehicle operations are constrained by limited spatial coverage and operational capabilities \cite{2025AUVreview}. Deploying multiple AUVs in cooperative formations enhances mission efficiency and robustness \cite{multiAUV1-JFR, multiAUV2-JOE, multiAUV3-TASE, multiAUV4-JOE}. Leader-follower formation control is a widely adopted cooperative paradigm \cite{leaderfollower1-JFR, leaderfollower2-JOE, leaderfollower3-Tsys}, in which a designated leader performs the primary mission while one or more followers maintain a prescribed relative pose, enabling collaborative mapping, inspection, and 3D reconstruction \cite{he2025review}. Reliable formation keeping depends critically on the follower's ability to estimate the leader's relative position and attitude accurately and in real time under challenging underwater sensing conditions \cite{AUVSurvey-CST}. Fig.~\ref{IntroFig} illustrates the proposed multimodal fusion framework, in which  the follower AUV acquires complementary visual, acoustic-ranging, and hydrodynamic measurements using a dedicated sensor suite and fuses them for real-time relative-pose estimation.

Unlike terrestrial robots that can rely on GPS, Wi-Fi, or LiDAR \cite{RAL-Wireless, TRO-Lidar}, AUVs face unique sensing and communication constraints. Radio-frequency signals are strongly attenuated in water, rendering GPS unavailable and severely limiting long-range wireless communication during submerged operation \cite{GPS-TVT}. Followers must therefore rely predominantly on onboard  sensing for relative pose estimation. Underwater imagery is further degraded by light absorption, scattering, color distortion, and suspended particulates, which severely reduce the reliability of optical measurements \cite{Vision-TPAMI}. Acoustic positioning systems such as LBL and USBL require external infrastructure, such as seafloor transponders or surface support vessels, which limits their applicability in ad hoc or wide-area missions \cite{usbl-jmse}. These perceptual challenges are compounded by the stringent computational and energy constraints under which AUVs operate \cite{AUVresource-jmse, AUVresource-TCE}. Designed for extended endurance, AUVs typically rely on low-power embedded processors with substantially less computational capacity than the server-grade GPUs commonly used to develop and evaluate terrestrial robotic systems \cite{AUVendurance-jmse, AUVendurance-JOE}. These constraints limit the onboard deployment of computationally intensive state-of-the-art algorithms \cite{TASELWMarineSurvey, LW-OE}.

A promising approach is to fuse complementary, low-cost sensors \cite{SensorFusion2}. A monocular camera provides high-resolution texture and geometric information at close range, but its performance deteriorates with distance and turbidity \cite{2021Vision-TIM}. An array of ultrasonic sensors provides robust, direct range measurements over longer distances, albeit with limited spatial resolution \cite{2022Acoustic-Sensor}. A pressure sensor array, employed as an artificial lateral line, detects the hydrodynamic wake generated by the leader's propulsion system, providing a persistent, low-power cue to the leader's presence and relative motion even when visibility is poor \cite{2025lateral-Tmech}. These three modalities, i.e., visual, acoustic, and hydrodynamic, are complementary, with each operating effectively over a different range of relative distances and under varying environmental conditions \cite{difsensing-sensors}. Exploiting this complementarity requires the effective integration of these heterogeneous, multiscale data streams \cite{optical-acoustic-RAL}.

Tri-modal optical--acoustic--pressure fusion has been shown to be feasible for static planar leader localization using a fixed sensing module and a manually positioned target \cite{Fusion-TMech}. The present work addresses the substantially different problem of continuous 6-DoF relative-pose tracking between two moving underwater vehicles. Such a dynamic setting introduces new challenges arising from viewpoint changes, coupled translational and rotational motion, time-varying modality reliability, temporal synchronization, and embedded closed-loop execution.

In this work, the follower's task is to continuously estimate the time-varying 6-DoF relative pose of the leader, expressed in the follower's body frame, from visual, acoustic, and hydrodynamic measurement streams. Because both vehicles move continuously, viewpoints, wake fields, and sensing conditions change continuously over time, requiring the three-dimensional translation and orientation to be estimated jointly in real time. Moreover, the three modalities differ substantially in dimensionality, sampling rate, and temporal characteristics, while their reliability varies with the relative motion of the vehicles and the surrounding environment. The fusion process must therefore adapt the contribution of each modality to its instantaneous reliability while operating within the computational constraints of embedded hardware.

However, many conventional fusion strategies face limitations when applied to this task under stringent computational constraints. Early fusion of raw data is computationally demanding and sensitive to modality-specific noise. Late fusion, while modular, often requires separate processing branches and may inadequately capture cross-modal interactions. Simple intermediate-fusion techniques based on static concatenation or summation also remain largely heuristic \cite{zhao2025multimodal}. They do not explicitly prioritize informative features or suppress redundant or uncertain data, leading to inefficient use of limited sensing and processing capacity.

The central challenge is therefore to integrate the three modalities effectively under time-varying sensing reliability: visual observations are high-dimensional and informative but susceptible to degradation; acoustic ranges are sparse, geometry-dependent, and susceptible to multipath effects; and hydrodynamic wakes are observable only locally, evolve over time, and depend on relative motion. Accordingly, effective multimodal fusion for underwater relative-pose tracking requires a heterogeneity-aware and reliability-adaptive design. Each modality should be encoded according to its own physical and temporal structure, and the fusion process should continuously adjust each modality's contribution according to its instantaneous measurement quality. Within this framework, information theory serves as a local design tool. A variational information bottleneck guides the compression of the high-dimensional visual stream
, and a KL-divergence regularizer shapes the distribution of learned hydrodynamic features
.

\subsection{Contributions}

In this paper, we present a reliability-adaptive multimodal perception framework for continuous 6-DoF relative-pose tracking between fully mobile underwater robots. The framework comprises three modality-specific encoders and a reliability-adaptive temporal fusion module: the Information Bottleneck Vision Encoder (IB-VE), the Physics-Aware Gated Acoustic Temporal Encoder (PA-GATE), the KL-Divergence Wake Encoder (KLD-WE), and the Adaptive Cross-Modal Fusion Module.

The IB-VE distills a compact, pose-relevant representation from visual inputs by minimizing a variational upper bound on the information-bottleneck objective, thereby suppressing task-irrelevant optical variations. The PA-GATE processes sparse ultrasonic readings by combining a static, physics-informed attention matrix with a dynamic, data-driven attention mechanism to extract spatiotemporal features efficiently. The KLD-WE employs a CNN-BiLSTM backbone with a distributional regularizer that constrains the extracted feature distribution relative to learnable reference distributions. The Adaptive Cross-Modal Fusion Module leverages a lightweight Adaptive Weight Generator (AWG) to assign modality weights based on learned confidence scores and a selective state-space model (Mamba-SSM) for efficient spatiotemporal integration, providing linear-time sequence modeling suitable for execution on embedded hardware. 

Extensive pool experiments demonstrate that the proposed framework achieves comparable or better pose-estimation accuracy than the evaluated fusion baselines. When deployed on an NVIDIA Jetson AGX Orin (64 GB), the second-stage fusion network achieves an inference latency of 17.62~ms, corresponding to a more than 2.6-fold speedup over Transformer-based fusion. This computational efficiency supports real-time closed-loop control on resource-constrained AUV hardware.

\begin{figure*}[thpb]
 \centering
 \includegraphics[width=1\textwidth]{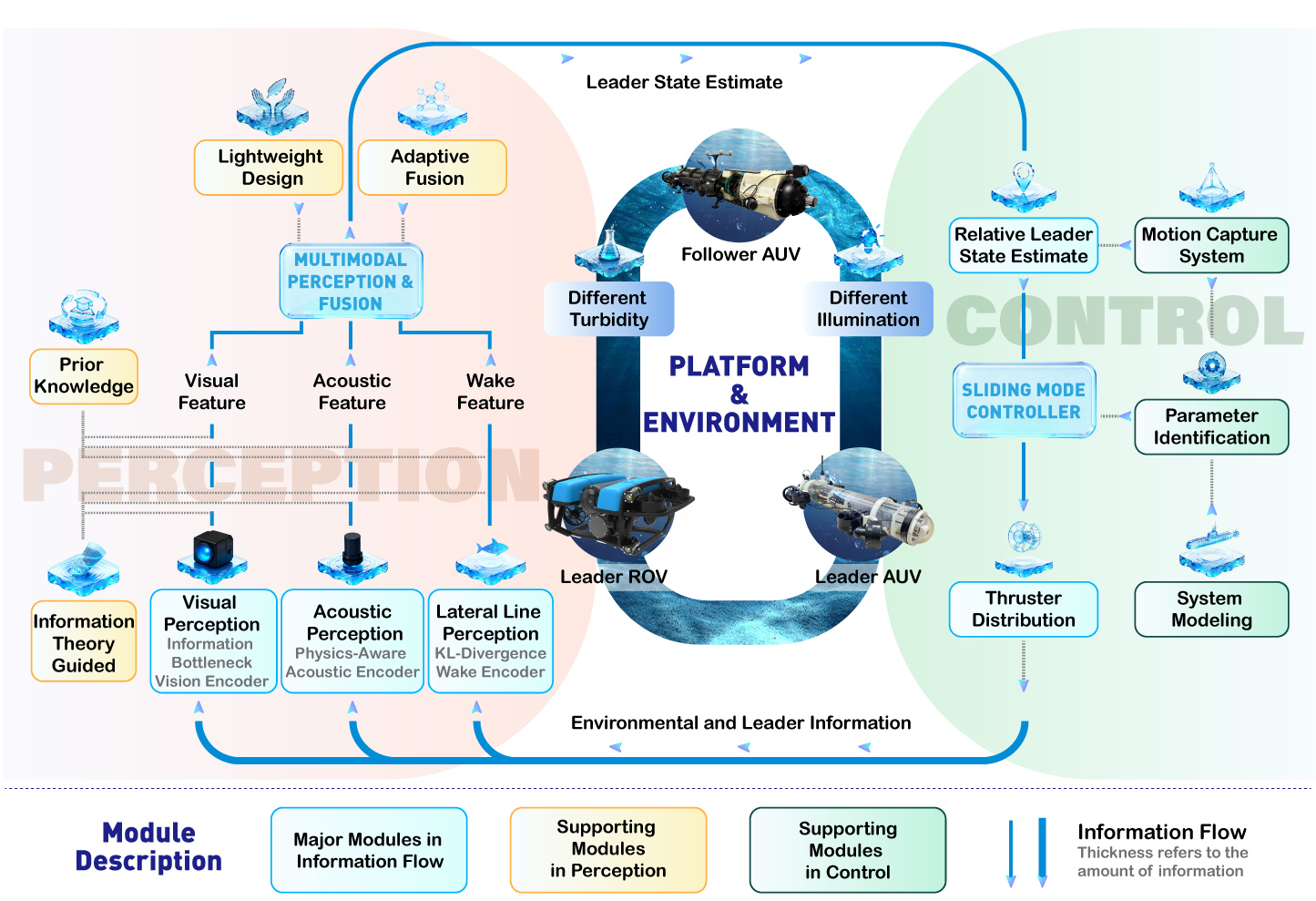}
 \caption{Overview of the reliability-adaptive multimodal perception and control loop for the leader-follower system. The system combines information-bottleneck visual encoding via IB-VE, physics-aware acoustic temporal encoding via PA-GATE, and hydrodynamic wake encoding with distributional regularization via KLD-WE to estimate the leader's relative pose. The modality-specific encoders incorporate relevant physical and statistical structure, while the fusion module adjusts modality contributions according to their estimated reliability.  The fused relative-pose estimate is provided to a sliding mode controller, supporting real-time formation control on resource-constrained AUV hardware.}
\label{Overview}
\end{figure*}

In summary, the key contributions of this work are threefold. First, we formulate and experimentally investigate dynamic 6-DoF leader-follower relative-pose estimation using complementary visual, acoustic-ranging, and hydrodynamic measurements, in a setting where both vehicles move continuously, producing time-varying viewpoints, wake fields, and modality reliability. Second, we develop a heterogeneity-aware fusion framework comprising modality-specific encoders and reliability-adaptive temporal fusion. The framework compresses visual representations, incorporates sensor geometry into the processing of sparse acoustic range measurements, models temporal wake dynamics from pressure sequences, and adaptively integrates the resulting representations using a selective state-space model. Third, we implement the complete perception pipeline on an onboard embedded computer and collect a multimodal dataset through 648 pool trials spanning two leader vehicle types, three illumination conditions, and nine trajectory patterns. Offline tests and closed-loop leader–follower experiments demonstrate continuous relative-pose estimation and formation maintenance under onboard computational constraints. The resulting multimodal dataset, together with an annotated underwater vehicle segmentation dataset, will be made publicly available. An overview of the perception and control loop is illustrated in Fig.~\ref{Overview}.

\section{Related Work}
\subsection{Multimodal Sensor Fusion in Underwater Environments}

Multimodal sensing addresses the limitations of individual underwater sensors. Cameras provide detailed close-range observations but degrade in turbid water, whereas acoustic sensors such as sonar and DVL offer greater robustness with lower spatial resolution and sparser measurements. Earlier localization methods used filtering \cite{2020RAL-fusion} and optimization \cite{2012JOE-Fusion}. Recent learning-based work includes cross-modal attention for paired acousto-optic detection \cite{2025JOE-Fusion}, attribute transfer and local descriptors for acousto-optic matching \cite{2021Sensors-Fusion}, deformable alignment and lightweight cross-attention \cite{2025JFR-Fusion}, and acoustic-optical fusion with enhancement and an interacting multiple-model adaptive Kalman filter \cite{2025Tmech-Fusion}.

Pressure arrays provide a low-power hydrodynamic cue complementary to vision and acoustics, particularly in low visibility \cite{2025TASE-Pressure, 2024OE-Pressure}. Tri-modal optical--acoustic--pressure fusion has also been explored for underwater leader localization \cite{Fusion-TMech}. These studies establish the value of heterogeneous sensing, but reliable dynamic 6-DoF tracking still requires the system to account for changes in each modality's reliability and the computational cost of processing asynchronous sensor streams.

\subsection{Efficient Learning for Underwater Perception}

AUV perception must balance accuracy with limited onboard computation. Existing approaches improve efficiency through knowledge distillation for underwater acoustic localization \cite{2025JOE-KnowledgeDistill}, decoupled distillation for object detection \cite{2025TGRS-KnowledgeDistill}, domain adaptation and few-shot learning for limited acoustic data \cite{2025JOE-KnowledgeDistill2}, and mixed precision, pruning, and quantization for ship detection \cite{2025SciRep-KnowledgeDistill}. These methods mainly optimize individual networks, rather than the joint requirements of heterogeneous sensing, changing modality quality, and temporal fusion on embedded robotic platforms.

Information-theoretic learning offers a complementary route to efficient perception by retaining task-relevant information while suppressing redundant features. It has been applied to bimanual imitation from RGB video \cite{2025RAL-robotlearning}, task-structure inference from demonstrations \cite{2025TRO-robotlearning}, f-divergence-based imitation learning \cite{2024IJRR-robotlearning}, and decentralized multi-robot navigation with cooperative perception \cite{2025TASE-robotylearning}. The Information Bottleneck (IB) principle supports compact representations, and the Variational IB makes this objective tractable for deep networks. In our framework, IB regularization compresses visual features in IB-VE, while KL divergence regularizes hydrodynamic features in KLD-WE. The acoustic encoder instead incorporates physical and geometric constraints. These modality-specific designs are integrated with reliability-adaptive weighting and lightweight state-space fusion to support dynamic 6-DoF tracking within embedded computational limits.

\section{System Overview}

This section describes the hardware configuration and system architecture of the open-source AUV \cite{OpenAUV}. The design emphasizes modularity, computational efficiency, and robust sensing for autonomous underwater operation.

\subsection{AUV Platform}

\begin{figure*}[thpb]
\begin{minipage}[t]{0.47\textwidth}
    \subfigure[Follower OpenAUV (top view).]{
    \label{AUV_topview}
    \includegraphics[width=\textwidth]{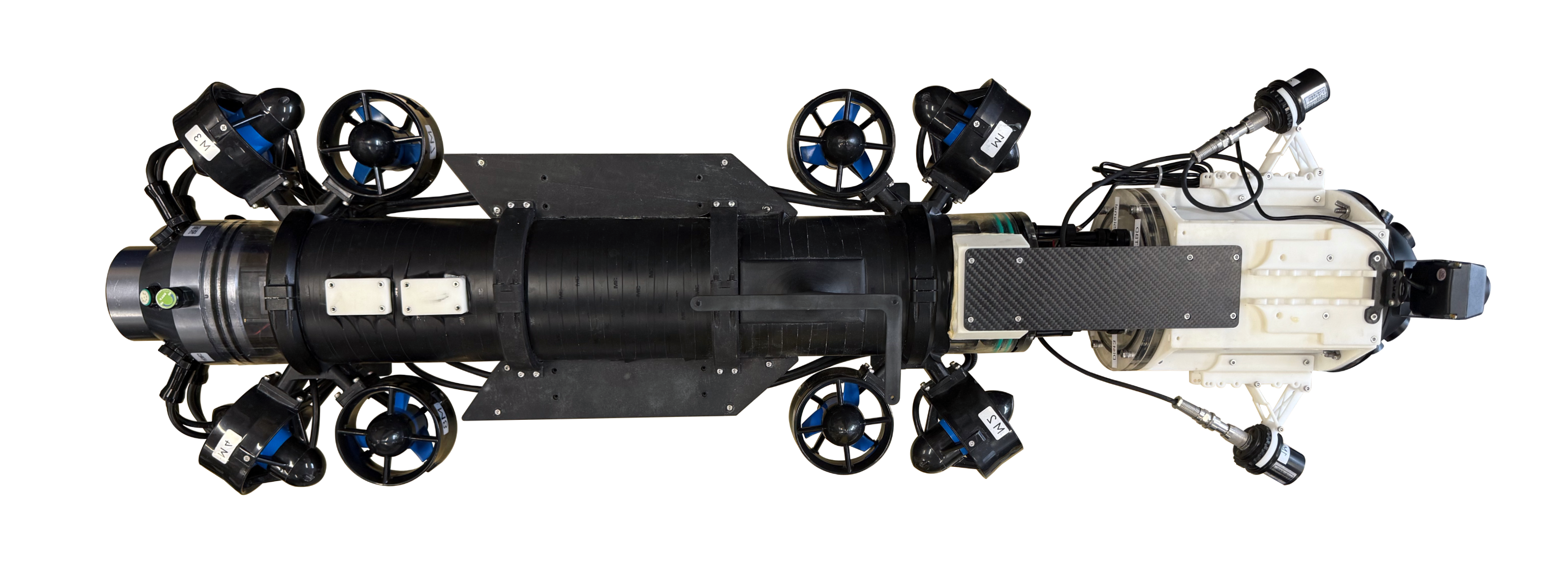}}
    \vspace{5mm}
    \subfigure[Follower OpenAUV (side view).]{
    \label{AUV_sideview}
    \includegraphics[width=\textwidth]{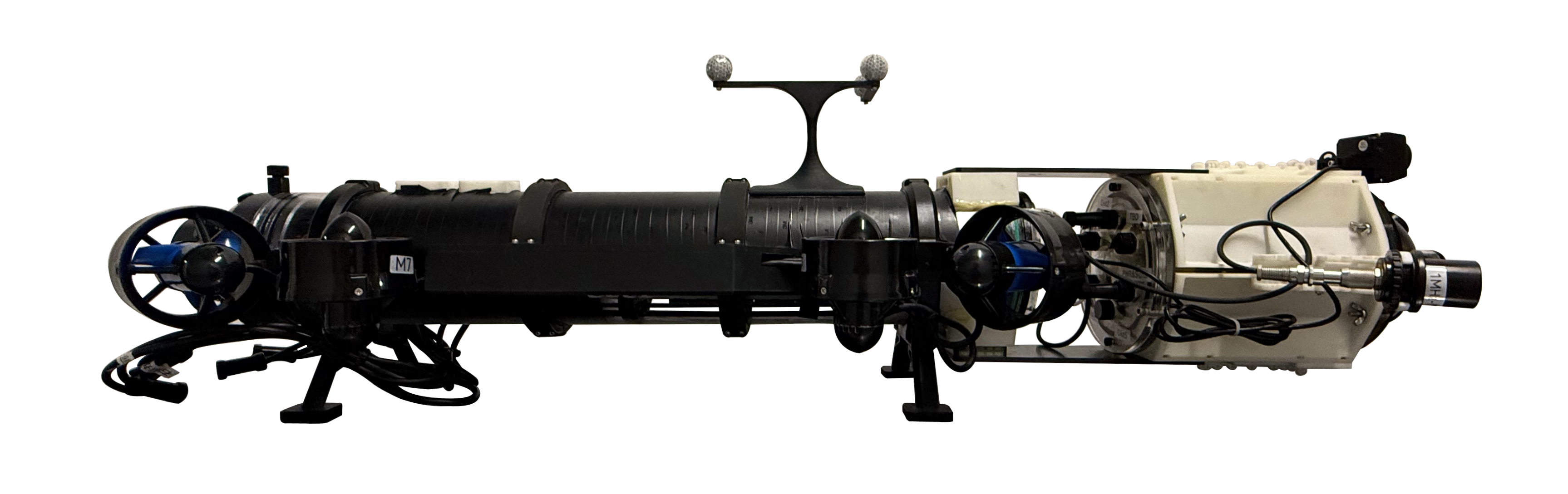}}
\end{minipage}
\hspace{10mm}
\begin{minipage}[t]{0.413\textwidth}
    \vspace{0pt}
    \centering 
    \subfigure[Sensing module (front view).]{
    \label{sensing_head}
    \includegraphics[width=\textwidth]{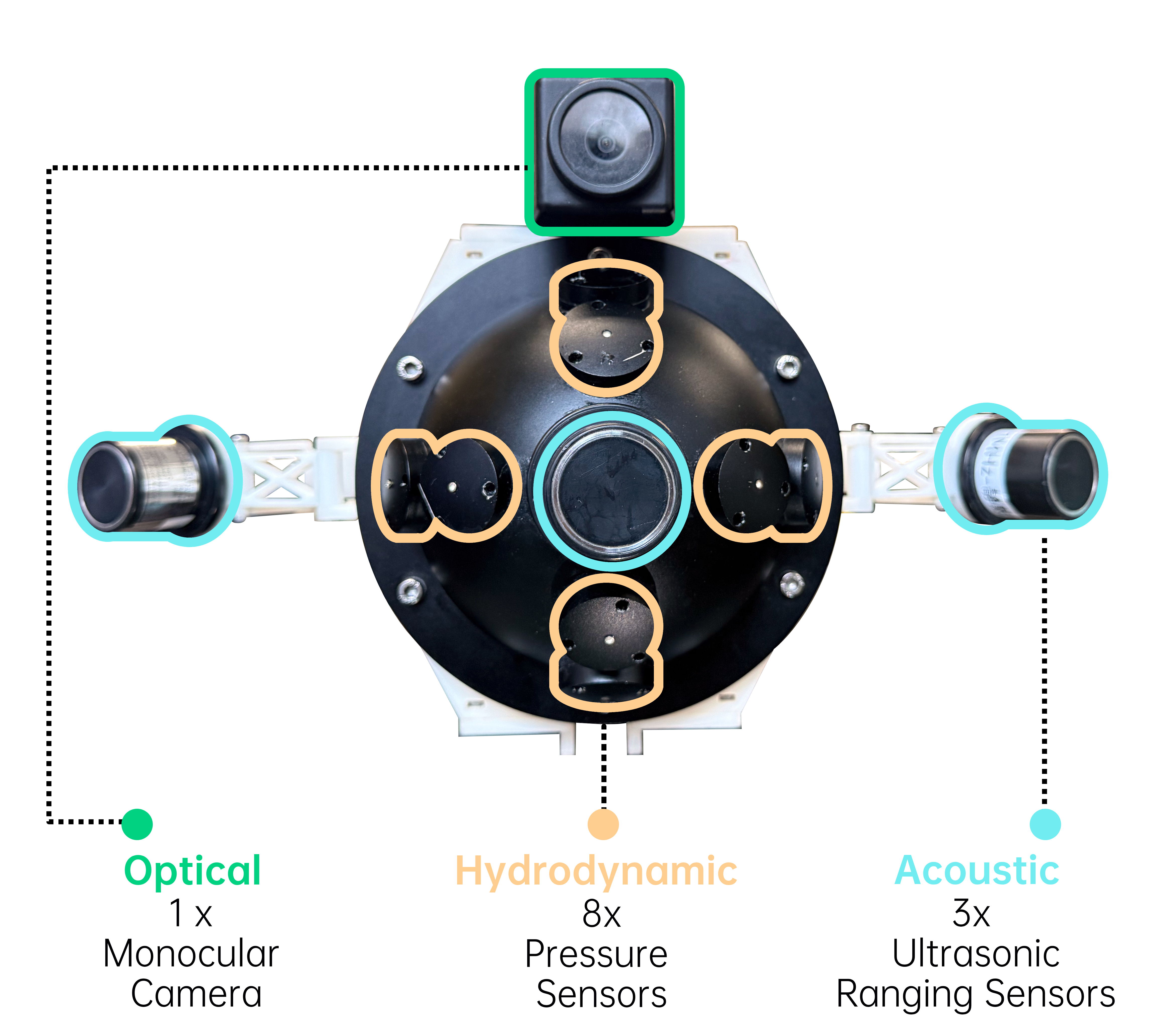}}
\end{minipage}
\centering
\caption{Hardware configuration of the follower AUV platform. The top view and side view of the follower OpenAUV are shown in (a) and (b), respectively. The follower OpenAUV consists of a rear main body and a forward sensing module, connected by two carbon-fiber plates. The main body is based on the same platform design as the leader OpenAUV, with minor modifications to its mechanical configuration. The front view of the sensing module in (c) shows the three sensing modalities. A monocular camera, three acoustic ranging sensors, and eight pressure sensors are outlined by green, blue, and orange contours, respectively.}
\label{Platforms}
\end{figure*}

The platform is a 1-m torpedo-shaped AUV with eight horizontal and vertical thrusters, providing control authority in surge, sway, heave, pitch, and yaw. The follower is an OpenAUV equipped with the sensing module shown in Fig.~\ref{AUV_topview} and Fig.~\ref{AUV_sideview}; the leader is either an unmodified OpenAUV or a BlueROV.

The sensing module (Fig.~\ref{sensing_head}) combines a waterproof 1920$\times$1080 RGB camera, three acoustic ranging sensors, and eight waterproof MS5837-02BA pressure sensors. The forward-facing acoustic sensor operates at 300 kHz with a wider field of view; two side-facing sensors operate at 1 MHz with 60$^\circ$ beam angles. Their different frequencies reduce mutual interference. The pressure array measures fluctuations induced by the leader's propeller wake.

The onboard computer is an NVIDIA Jetson AGX Orin module, rated at up to 275 TOPS of AI performance. This module supports real-time multimodal perception and closed-loop control on board the follower AUV.

\subsection{Physical Environment}

\begin{figure}[thpb]
 \centering
 \subfigure[MoCap system (top view).]{
\label{mocap_topview}
\includegraphics[width=0.25\textwidth]{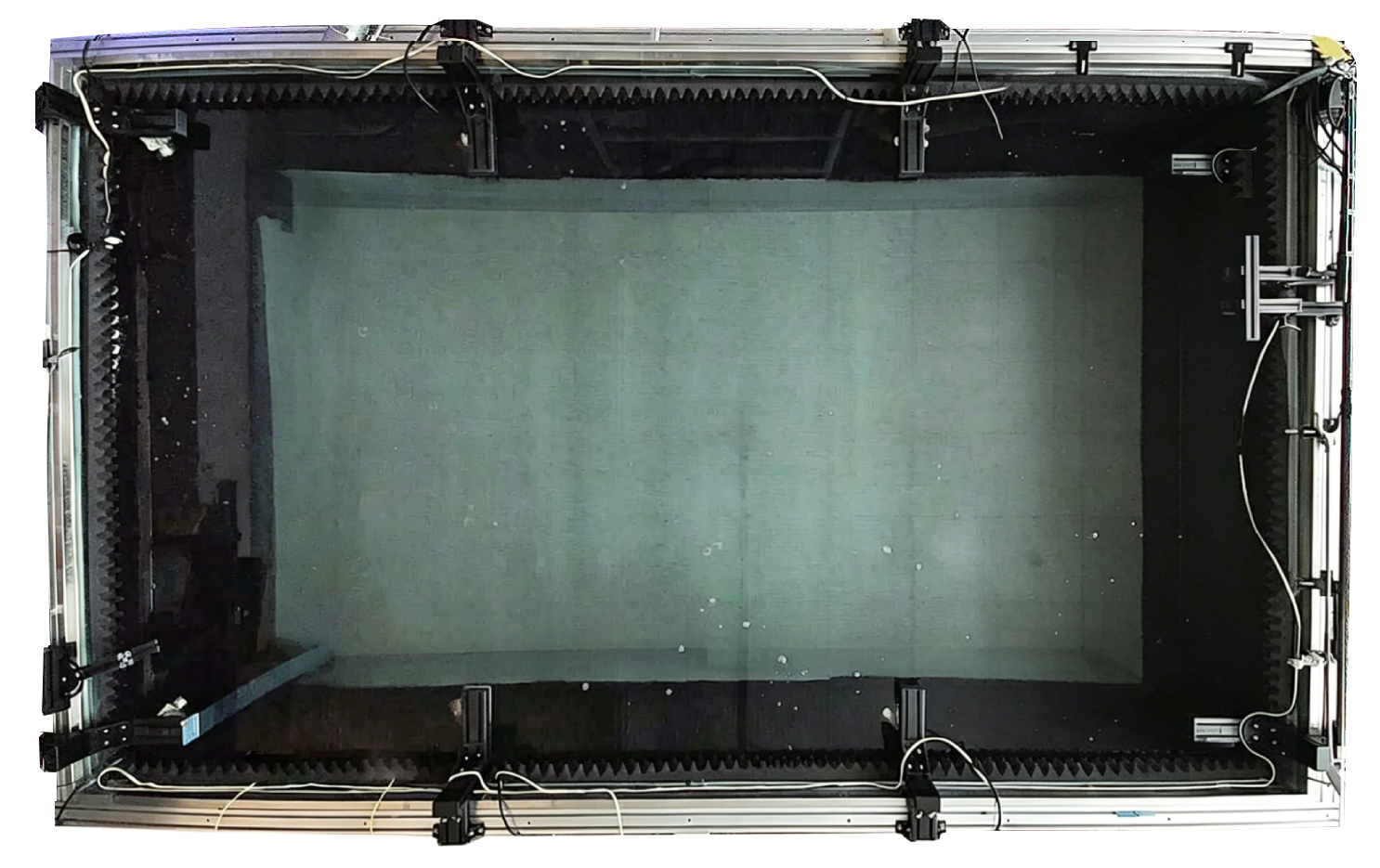}}
\subfigure[MoCap system (side view).]{
\label{mocap_sideview}
\includegraphics[width=0.215\textwidth]{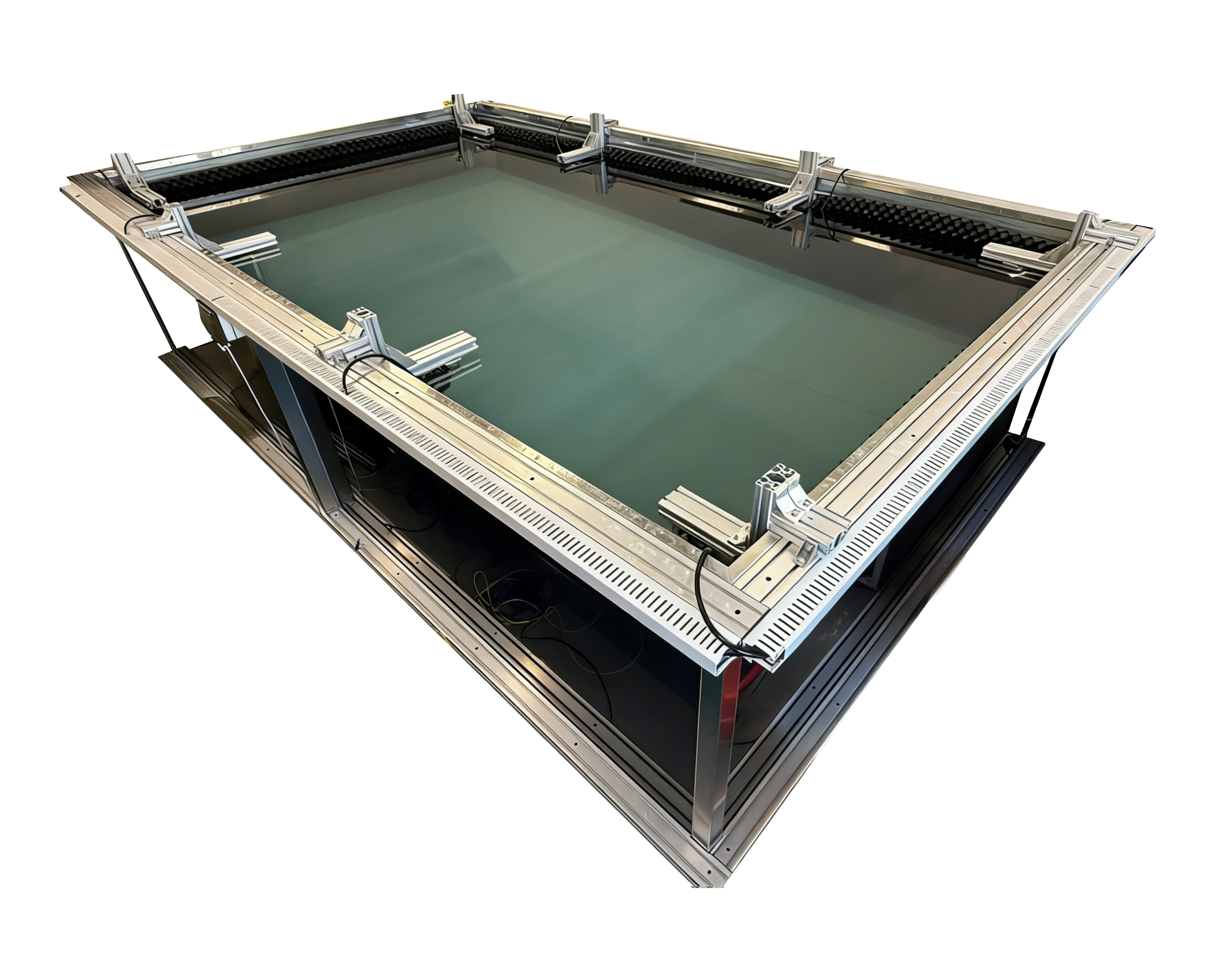}}
\caption{Experimental pool and underwater motion-capture system. The system comprises eight high-resolution underwater optical cameras, with four positioned along each long side of the pool.}
\label{Pool w/ Mocap}
\end{figure}

Experiments were conducted in a 4~m $\times$ 2~m pool with an underwater Nokov MoCap system comprising eight Mars1.3HW UW cameras (1280$\times$1024, 240~Hz). The system provides 6-DoF ground truth with reported positional accuracy of $\pm$3~mm; its layout is shown in Fig.~\ref{mocap_topview} and Fig.~\ref{mocap_sideview}.

\section{Perception Framework}

\begin{figure*}[thpb]
 \centering
 \includegraphics[width=1\textwidth]{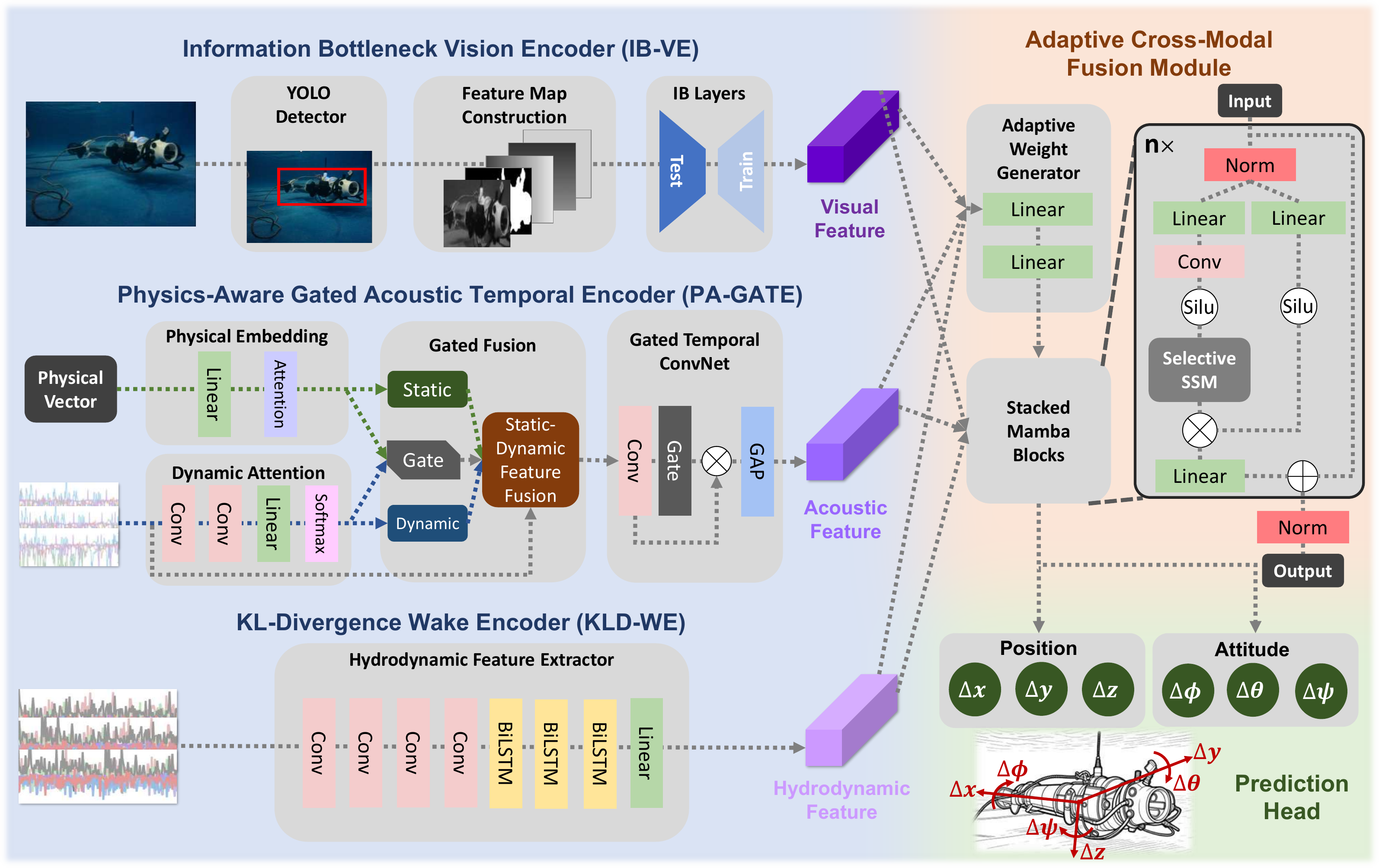}
 \caption{Overview of the proposed perception framework. The perception framework incorporates three encoder modules, one adaptive fusion module and one regression head. The encoder, fusion, and regression stages are indicated by blue, orange, and green backgrounds, respectively. A YOLO-based front end first detects and segments the leader, from which a task-specific five-channel feature map is constructed. The Information Bottleneck Vision Encoder (IB-VE) then extracts a compact visual representation.  The Physics-Aware Gated Acoustic Temporal Encoder (PA-GATE) processes static sensor-geometry descriptors and time sequences of acoustic range measurements to obtain the acoustic representation. The KL-Divergence Wake Encoder (KLD-WE) encodes pressure sequences acquired by the distributed artificial lateral line using lightweight convolutional and BiLSTM layers to obtain the hydrodynamic representation. The features from the three sensing modalities are adaptively fused by the Adaptive Cross-Modal Fusion Module, which combines a Mamba-based selective state-space model with the Adaptive Weight Generator (AWG). The prediction head maps the fused representation to the leader's 6-DoF pose relative to the follower.}
\label{Alg-Overview}
\end{figure*}

The framework comprises vision, acoustic, and hydrodynamic encoders, an adaptive fusion module, and a regression head. The encoders process images, acoustic ranges, and pressure sequences, respectively, and the fusion module estimates the leader's relative 6-DoF pose. Fig.~\ref{Alg-Overview} summarizes the architecture.

\subsection{Vision Encoder}
\label{sec:vision_encoder}

This subsection presents the Information Bottleneck Vision Encoder (IB-VE), a neural architecture designed to extract compact visual features for underwater leader–follower tracking. The IB-VE first constructs a task-specific feature map from detection and segmentation results and spatial priors. It then applies the Information Bottleneck (IB) principle to learn a compact stochastic representation for subsequent multimodal fusion.

\subsubsection{Lightweight Visual Feature Construction}

Underwater relative pose estimation relies on reliable visual cues, yet optical degradation and computational constraints pose significant challenges. To address these issues, we design a pipeline that integrates detection and segmentation results from a YOLO-Seg model to generate a compact $128\times 128\times 5$ feature map suitable for real-time pose estimation on embedded hardware.

\textbf{YOLO Detector} A lightweight YOLOv11n-seg model serves as the front-end detector and segmenter. The model is fine-tuned on a small annotated underwater image dataset for leader localization and segmentation.

\textbf{Feature Map Construction with Prior Knowledge} The final $128\times128\times5$ feature map stacks task-critical visual cues, with each channel designed to provide information relevant to relative-pose estimation.

The first channel is a water-condition-adaptive intensity map formed by weighting the RGB channels according to the output of a water-condition classifier. For highly turbid conditions, we use $0.1R+0.7G+0.2B$, assigning the greatest weight to the less attenuated green channel. For clear water, we use $0.3R+0.5G+0.2B$. This single channel preserves intensity and texture cues useful for attitude estimation while reducing the influence of more strongly attenuated color channels.

The second channel is a binary segmentation mask that isolates the leader from the background, allowing the network to focus on its silhouette and spatial extent. The target centroid, shape, and aspect ratio provide useful cues for relative-position and attitude estimation.

The third and fourth channels are global coordinate-encoding maps calculated as
\begin{equation}
    pos_x(i,j)=x^\text{ul}_\text{n}+(j/f_\text{dw}) \times w_\text{n},
\end{equation}
\begin{equation}
    pos_y(i,j)=y^\text{ul}_\text{n}+(i/f_\text{dh}) \times h_\text{n},
\end{equation}
where $w_\text{n}$ and $h_\text{n}$ are normalized bounding box dimensions, $(x^\text{ul}_\text{n}, y^\text{ul}_\text{n})$ is the normalized coordinates of the upper-left corner of the bounding box, and $f_\text{dw}$ and $f_\text{dh}$ are the width and height of the constructed feature map, respectively. These channels encode the leader AUV's absolute image position and apparent scale in the original image, providing cues related to relative displacement and distance-dependent size variations.

The fifth channel spatially replicates the YOLO confidence score, providing the downstream network with a cue to the reliability of the visual detection. This cue allows the fusion module to learn how the visual contribution should vary with detection confidence.

The five channels retain appearance, silhouette, spatial, and detection-confidence cues while reducing the representation passed to the IB layers.

\subsubsection{Feature Compression Guided by Information Theory}

The IB principle \cite{tishby2000IB}, implemented through a variational approximation \cite{alemi2016VIB, shwartz2017opening}, encourages compact representations that suppress task-irrelevant image variation. Here, the decoder reconstructs the input feature map ($Y=X$), making the reconstruction term a KL-regularized variational autoencoding objective. Downstream pose supervision further aligns the compressed features with relative-pose estimation (Section~\ref{sec:loss_design}).

\textbf{Variational information bottleneck}

Let $X$ be the input feature map and $Z$ its stochastic latent representation. The encoder limits retained information while preserving reconstruction, using a variational distribution $q(z)$ to approximate the latent marginal and a decoder $r(y|z)$ to reconstruct $X$. The resulting objective is an upper bound on the information-bottleneck objective,
\begin{equation}
    \mathcal{J}_{\mathrm{IB}} =
    \beta\,\mathbb{E}_{x}\!\left[\mathrm{KL}\bigl(p(z|x)\|q(z)\bigr)\right]
    - \mathbb{E}_{x,z}\!\left[\log r(x|z)\right].
\end{equation}
Here, $\beta$ controls the trade-off between latent compression and reconstruction fidelity. The latent variable is sampled with the reparameterization trick, $z=\boldsymbol{\mu}(x)+\boldsymbol{\sigma}(x)\odot\boldsymbol{\epsilon}$, where $\boldsymbol{\epsilon}\sim\mathcal{N}(0,I)$. This formulation provides a tractable training objective without requiring explicit estimation of the latent marginal distribution.

\subsubsection{Architecture Design}

The normalized $128\times128\times5$ feature map passes through a lightweight convolutional encoder and stochastic bottleneck (Table~\ref{IB-VE}). During training, the decoder reconstructs the input and the KL term regularizes the latent distribution. At inference, the decoder is removed and $z=\mu$ is used to produce a 512-dimensional feature $(D_\text{v}=512)$. Italicized table entries are training-only.

\begin{table}[thpb]
\caption{Architecture of IB-VE}
\centering
    \renewcommand\arraystretch{1.4}
    \begin{tabularx}{0.48\textwidth}{ccccc}
        \toprule
        Stage & Layer & Output Size & Operator & \#Out\\
        \midrule
        \multirow{2}*{Input} & Input Map & 128 $\times$ 128 & YOLO-SEG & 5 \\
        & CoordConv & 128 $\times$ 128 & Concat $(x,y)$ & 7 \\
        \midrule
        \multirow{6}*{Encoder}
        & Conv 1 & 64 $\times$ 64 & Conv(3,3) & 16 \\
        & Conv 2 & 32 $\times$ 32 & Conv(3,3) & 32 \\ 
        & Conv 3 & 16 $\times$ 16 & Conv(3,3) & 64 \\
        & Conv 4 &  8 $\times$  8 & Conv(3,3) & 128 \\
        & Conv 5 &  4 $\times$  4 & Conv(3,3) & 256 \\
        & Conv 6 &  2 $\times$  2 & Conv(3,3) & 512 \\
        \midrule
        \multirow{5}*{Bottleneck}
        & Flatten & 2048 & Vectorize & NA \\
        & Linear $\mu$ & 512 & FC & NA \\
        & Linear $\sigma$ & 512 & FC & NA \\
        & \textit{Train Sample} & 512 & $\bm{z}=\bm{\mu}+\bm{\epsilon}\odot\bm{\sigma}$ & NA \\
        & Test Sample & 512 & $\bm{z}=\bm{\mu}$ & NA \\
        \midrule
        \multirow{8}*{\textit{Decoder}}
        & \textit{Linear} & 2048 & \textit{FC} & NA \\
        & \textit{Unflatten} & 2 $\times$  2 & \textit{Reshape} & 512 \\
        & \textit{DeConv 1} &  4 $\times$  4 & \textit{TransConv(3,3)} & 256 \\ 
        & \textit{DeConv 2} &  8 $\times$  8 & \textit{TransConv(3,3)} & 128 \\
        & \textit{DeConv 3} & 16 $\times$ 16 & \textit{TransConv(3,3)} &  64 \\
        & \textit{DeConv 4} & 32 $\times$ 32 & \textit{TransConv(3,3)} &  32 \\
        & \textit{DeConv 5} & 64 $\times$ 64 & \textit{TransConv(3,3)} &  16 \\
        & \textit{DeConv 6} & 128 $\times$ 128 & \textit{TransConv(3,3)} & 5 \\
        \bottomrule
    \end{tabularx}
    \begin{tablenotes}
    \footnotesize
    \item The stages, layers, and operators in \textit{italics} are training-only.
    \end{tablenotes}
\label{IB-VE}
\end{table}

\subsubsection{Loss Function Design}

The IB-VE loss $\mathcal{L}_\text{VIB}$ comprises a reconstruction loss and a KL divergence loss, $\mathcal{L}_\text{VIB}=\mathcal{L}_\text{reg}+\mathcal{L}_\text{KL}$.

We model the input feature map, which serves as the reconstruction target $Y$, using a Gaussian distribution conditioned on the latent representation $Z$,
\begin{equation}
    r(\bm{y}|\bm{z}) = \mathcal{N}(\bm{y}|\bm{\mu}(\bm{z}),\bm{\sigma}^2(\bm{z})),
\end{equation}
where $\bm{\mu}(\bm{z})$ is the mean and $\bm{\sigma}^2(\bm{z})$ is the observation variance.

The resulting variational objective is
\begin{equation}
\begin{split}
    \mathcal{L}_\text{VIB} = & \frac{1}{N} \sum_{n=1}^{N} \{ \beta \times \text{KL} [p(Z|x_n),q(Z)] \\
& + \frac{1}{M} \sum_{m=1}^M [ - \log r(y_n|f(x_n,\epsilon_m))] \}.
\end{split}
\end{equation}

The KL divergence loss $\mathcal{L}_\text{KL}$ constrains the latent distribution. With a fixed spherical Gaussian prior $q(Z)=\mathcal{N}(z|\bm{0},\bm{I})$, the loss is
\begin{equation}
    \mathcal{L}_\text{KL} = \frac{1}{N} \sum_{n=1}^{N} \beta \times [ \frac{1}{2}(\text{tr}(\bm \sigma_x^2)+||\bm\mu_x||^2 - d_z - \log |\bm\sigma_x^2|)].
\end{equation}

With the decoder predicting the mean and log variance of the input feature map, $[\bm\mu(\bm{z}),\log \bm{\sigma}^2(\bm{z})]=f_d(\bm{z})$, the reconstruction loss is
\begin{equation}
    \mathcal{L}_\text{reg} = \frac{1}{NM} \sum_{n=1}^{N} \sum_{m=1}^{M} \left[ \frac{1}{2} \log (2 \pi \sigma^2(z)) + \frac{(y_n-\mu(z))^2}{2\sigma^2(z)} \right].
\end{equation}

The reconstruction term provides self-supervision, while the downstream fusion and regression losses supervise relative-pose estimation.

\subsection{Acoustic Encoder}

To process sparse and noisy readings from multiple acoustic range sensors, we propose the Physics-Aware Gated Acoustic Temporal Encoder (PA-GATE). PA-GATE combines geometry-conditioned attention, data-driven attention, and gated temporal convolution to encode the acoustic measurement sequences.

\subsubsection{Physics-Aware Sensor Embedding}

A Physics-Aware Spatial Encoding layer computes a static attention matrix based on learnable embeddings of the sensors' physical properties. For the $i$-th sensor, a compact physical feature vector $\bm{p}_i \in \mathbb{R}^{d_\text{physical}}$ comprises its 3D Cartesian coordinates $\bm{c}_i$ relative to the AUV's center of mass, its orientation $\bm{o}_i$ represented by a unit vector, and its characteristic beam width $\alpha_i$, denoted as $\bm{p}_i = [\bm{c}_i, \bm{o}_i, \alpha_i]^\text{T}$, with $d_\text{physical}=7$.

This descriptor is projected into the model feature space via
\begin{equation}
    \bm{e}_i = \text{GELU} (\bm{W}_\text{proj} \bm{p}_i + \bm{b}_\text{proj}),
\end{equation}
where $\bm{W}_\text{proj} \in \mathbb{R}^{d_\text{model} \times d_\text{physical}}$ and $\bm{b}_\text{proj} \in \mathbb{R}^{d_\text{model}}$ are trainable. The sensor embedding matrix $\bm{E}\in \mathbb{R}^{S \times d_\text{model}}$ is formed by stacking all $\bm{e}_i$.

The static spatial attention matrix is
\begin{equation}
        \bm{M}_{\text{s}} = \text{Softmax}\left( \frac{ (\bm{E}\bm{W}_q) (\bm{E}\bm{W}_k)^\text{T} }{ \sqrt{d_\text{model}} } \right),
\end{equation}
where $\bm{M}_{\text{s}}(i,j)$ denotes the geometry-conditioned attention assigned by sensor $i$ to sensor $j$.

The physical descriptors provide a geometry-conditioned inductive bias, while the projection and attention parameters are optimized end-to-end to capture task-relevant relationships among the sensors.

\subsubsection{Dynamic Data-Driven Attention}

A dynamic data-driven attention branch computes time-varying sensor weights from the acoustic measurement sequence $\bm{A}_\text{r}$. Lightweight 1D convolutional layers operate along the temporal dimension before computing sensor-wise attention,
\begin{equation}
    \bm{A}_\text{tf} = \text{Conv1D}(\text{Conv1D}(\bm{A}_\text{r})),
\end{equation}
\begin{equation}
    \bm{A}_\text{d} = \text{Softmax}( \text{Linear}(\bm{A}_\text{tf})),
\end{equation}
where $\bm{A}_\text{r}, \bm{A}_\text{tf}, \bm{A}_\text{d} \in \mathbb{R}^{B \times T \times S}$. GELU activation and LayerNorm are used within the branch to stabilize optimization.

\subsubsection{Gated Fusion Attention}

A gated fusion mechanism combines the static geometry-based and dynamic measurement-based reliability estimates. We compute a static reliability vector $\bar{m}_\text{s} \in \mathbb{R}^{S}$ by averaging the attention received by each sensor, and a dynamic reliability vector $\bar{a}_\text{d} \in \mathbb{R}^{B \times S}$ by averaging the measurement-dependent attention over the temporal window,
\begin{equation}
    \bar{m}_\text{s}(i) = \frac{1}{S} \sum_{j} M_\text{s}(j,i),
\end{equation}
\begin{equation}
    \bar{a}_\text{d} = \frac{1}{T} \sum_{t=1}^{T} A_\text{d}(t,i).
\end{equation}

These are concatenated and processed by an MLP with sigmoid activation to generate a gating weight $g \in \mathbb{R}^{B \times T \times S}$,
\begin{equation}
    g = \sigma(\text{MLP}(\text{Concat}(\bar{m}_\text{s}, \bar{a}_\text{d}))).
\end{equation}

The fused features are
\begin{equation}
    A_\text{f} = \bm{g} \odot (M_\text{s} A_\text{r}) + (\bm{1} - \bm{g}) \odot (A_\text{d} \odot A_\text{r}).
\end{equation}

This mechanism enables an adaptive balance between physical priors and time-varying measurements.

\subsubsection{Gated Temporal Convolution Network}

The fused features are processed by a lightweight gated temporal convolution network (TCN) with output dimension $D_\text{a}=128$,
\begin{equation}
    \bm{F} = \text{Conv1D}(\bm{A}_{\text{f}}),
\end{equation}
\begin{equation}
    \bm{G} = \sigma( \text{Conv1D}(\bm{A}_{\text{f}}) ),
\end{equation}
\begin{equation}
    \bm{Z}_{\text{tcn}} = \mathbf{G} \times \text{GELU}(\bm{F}),
\end{equation}
where $\mathbf{Z}_{\text{tcn}} \in \mathbb{R}^{B \times T \times D_\text{a}}$.

Global average pooling over the temporal dimension  aggregates the sequential features into a fixed-length acoustic feature vector,
\begin{equation}
    \bm{Z}_\text{a} = \text{GAP}(\bm{Z}_{\text{tcn}}).
\end{equation}

This design is parameter-efficient and parallelizable, supporting efficient inference on embedded hardware.

\subsection{Hydrodynamic Encoder}
\label{sec:wake_encoder}

This subsection presents the KL-Divergence Wake Encoder (KLD-WE), a neural architecture for extracting hydrodynamic features from pressure sensor data. By integrating distributional learning with bidirectional temporal processing, the encoder captures vehicle wake signatures while maintaining robustness against environmental variations.

\subsubsection{Architecture Design}

The core feature extraction module combines 1D CNN layers and a three-layer stacked bidirectional LSTM (BiLSTM) to capture temporal wake dynamics. This CNN-BiLSTM backbone follows the architecture in \cite{Fusion-TMech}; the component newly introduced in the present framework is the distributional regularization described in the loss design below. The input comprises 64-time-step sequences acquired from eight pressure sensors, and the output is a 256-dimensional feature vector formed by concatenating the forward and backward features. The KLD-WE architecture is detailed in Table~\ref{KLD-WE}. The extracted hydrodynamic feature dimension is $D_\text{p}=64$.

\begin{table}[thpb]
\caption{Architecture of KLD-WE}
\centering
    \renewcommand\arraystretch{1.4}
    \begin{tabular}{cccc}
        \toprule
        Stage & Output Size & Operator & \#Out\\
        \midrule
        \multirow{2}*{1} & 64 $\times$ 1 & Conv(3,1) & 128 \\ 
        & 64 $\times$ 1 & Conv(3,1) & 128 \\
        \midrule
        \multirow{2}*{2} & 64 $\times$ 1 & Conv(3,1) & 64 \\ 
        & 64 $\times$ 1 & Conv(3,1) & 64 \\
        \midrule
        3 & 1 $\times$ 1 & BiLSTM(128,3) & 256 \\
        \midrule
        4 & 1 $\times$ 1 & FC & 64 \\
        \bottomrule
    \end{tabular}
\label{KLD-WE}
\end{table}

\subsubsection{Loss Function Design}

The encoder incorporates learnable reference distribution parameters that are jointly optimized with the network to capture recurring statistical structure in the hydrodynamic features observed during training. The loss $\mathcal{L}_\text{P}$ measures the KL divergence between the extracted feature distribution and the reference distribution,
\begin{equation}
\begin{split}
    \mathcal{L}_\text{P} & = \text{KL} (P_\text{f}||P_\text{ref}) \\
& = \frac{1}{2} \left[ \log\frac{\sigma_\text{ref}^2}{\sigma_\text{f}^2} + \frac{\sigma_\text{f}^2+\left(\mu_\text{f}-\mu_\text{ref}\right)^2}{\sigma_\text{ref}^2} - 1\right].
\end{split}
\end{equation}

We emphasize that $\mathcal{L}_\text{P}$ serves as distributional regularization rather than a physically grounded information-theoretic objective: the reference distributions are learned from data and are not assigned an explicit physical interpretation. Their role is to encourage statistically regular hydrodynamic representations and reduce sensitivity to transient pressure fluctuations.

\subsection{Adaptive Cross-Modal Spatiotemporal Fusion}

At each fusion step, modality-specific multilayer perceptrons estimate the reliability of visual, acoustic, and hydrodynamic features. Softmax normalization produces modality weights $\alpha_v$, $\alpha_a$, and $\alpha_p$. The weighted features are concatenated and processed by a selective state-space model (Mamba-SSM), then a compact regression head estimates the leader's relative position and attitude.

\subsubsection{Adaptive Weight Generator}

The underwater environment causes significant fluctuations in modality reliability. Visual observations degrade under turbidity and poor illumination, while acoustic measurements are susceptible to multipath interference and geometry-dependent ranging uncertainty. Hydrodynamic measurements can also be affected by ambient flow disturbances and wake dissipation. A fixed fusion strategy cannot adapt to these time-varying sensing conditions.

The AWG dynamically assigns weights at the modality level. Features from each modality are processed by separate lightweight, modality-specific MLPs to generate scalar reliability scores. These scores are normalized via Softmax to produce weights $\alpha_v$, $\alpha_a$, and $\alpha_p$ that sum to one. These weights scale the encoded modality features before their concatenation and subsequent Mamba fusion.

\subsubsection{Cross-Modal Fusion Module}

The cross-modal fusion module employs a selective state-space model (Mamba-SSM) \cite{gu2024mamba} as its core architecture. At each fusion step, the weighted 512-dimensional visual, 128-dimensional acoustic, and 64-dimensional hydrodynamic features are concatenated into a 704-dimensional multimodal feature vector. A temporal sequence of these vectors is then provided to the Mamba-SSM. This sequence is processed through a single Mamba-SSM layer with a state dimension of 64 and an expansion factor of 2, and the resulting representation is projected to a 224-dimensional fused feature vector. Mamba-SSM has a computational complexity of $O(n)$, compared with the $O(n^2)$ complexity of standard self-attention in Transformers. This linear scaling facilitates efficient temporal modeling of long multimodal sequences on resource-constrained AUV hardware.

\subsection{Prediction Head}

The fused feature representation is mapped to the current relative position and attitude by a compact regression head. The head uses an MLP to decode the 224-dimensional fused feature vector into a 6-DoF relative pose vector $[\Delta x, \Delta y, \Delta z, \Delta \phi, \Delta \theta, \Delta \psi]$. The translation components are expressed in meters along the axes of the follower's body-fixed frame, and the rotation components are Euler angles in radians.

The MLP comprises two hidden layers with GELU activation and LayerNorm applied before the corresponding linear transformation. The final output layer produces normalized regression values. A combined loss $\mathcal{L}_\text{pos}+\mathcal{L}_\text{att}$ supervises training. The translation component uses the smooth L1 loss, which is less sensitive to outliers than MSE. The rotation component uses cosine similarity loss for yaw to account for angular periodicity and MSE for roll and pitch.

\subsection{Loss Design}
\label{sec:loss_design}

The total loss is
\begin{equation}
    \mathcal{L} = \lambda_1\mathcal{L}_\text{pos} + \lambda_2\mathcal{L}_\text{att} + \lambda_3\mathcal{L}_\text{VIB} + \lambda_4\mathcal{L}_\text{P},
\end{equation}
where $\mathcal{L}_\text{pos}$, $\mathcal{L}_\text{att}$, $\mathcal{L}_\text{VIB}$, and $\mathcal{L}_\text{P}$ are the position loss, attitude loss, IB-VE loss, and KLD-WE loss, respectively, with corresponding weighting coefficients $\lambda_1$ through $\lambda_4$.

\section{Modeling and Control Strategy}
This section covers the dynamic modeling of the rigid-body torpedo-shaped AUV, followed by the design of a sliding mode controller (SMC).

\subsection{AUV Modeling}

For each AUV $i\in\{\mathrm{L},\mathrm{F}\}$, where $\mathrm{L}$ and $\mathrm{F}$ denote the leader and follower, respectively, the 6-DoF kinematics and dynamics are given by~\cite{fossen2011handbook,shen2016integrated}
\begin{align}
    \dot{\boldsymbol{\eta}}_i
    &=\boldsymbol{J}(\boldsymbol{\eta}_i)\boldsymbol{\nu}_i, \\
    \boldsymbol{M}_i\dot{\boldsymbol{\nu}}_i
    +\boldsymbol{C}_i(\boldsymbol{\nu}_i)\boldsymbol{\nu}_i
    +\boldsymbol{D}_i(\boldsymbol{\nu}_i)\boldsymbol{\nu}_i
    +\boldsymbol{g}_i(\boldsymbol{\eta}_i)
    &=\boldsymbol{\tau}_i+\boldsymbol{\tau}_{d,i},
    \label{dynamic}
\end{align}
where $\boldsymbol{\eta}_i=[\boldsymbol{p}_i^\mathrm{T},
\boldsymbol{\Theta}_i^\mathrm{T}]^\mathrm{T}\in\mathbb{R}^6$
is the inertial-frame pose and
$\boldsymbol{\nu}_i=[\boldsymbol{v}_i^\mathrm{T},
\boldsymbol{\omega}_i^\mathrm{T}]^\mathrm{T}\in\mathbb{R}^6$
is the body-fixed velocity. The remaining terms represent inertia, Coriolis and centripetal effects, hydrodynamic damping, hydrostatic restoring forces, control input, and external disturbances, with their standard definitions.

The tracking state is the pose of the leader expressed in the follower frame:
\begin{equation}
    {}^{\mathrm{F}}\boldsymbol{T}_{\mathrm{L}}
    =
    \left({}^{\mathrm{W}}\boldsymbol{T}_{\mathrm{F}}\right)^{-1}
    {}^{\mathrm{W}}\boldsymbol{T}_{\mathrm{L}},
    \qquad
    \boldsymbol{\xi}
    =
    \begin{bmatrix}
        {}^{\mathrm{F}}\boldsymbol{p}_{\mathrm{L}}^\mathrm{T} &
        {}^{\mathrm{F}}\boldsymbol{\Theta}_{\mathrm{L}}^\mathrm{T}
    \end{bmatrix}^{\mathrm{T}}.
\end{equation}
Differentiating this relative pose and combining the dynamics of both vehicles yields the follower-actuated relative model
\begin{equation}
    \ddot{\boldsymbol{\xi}}
    =
    \boldsymbol{f}_r(\boldsymbol{\xi},\dot{\boldsymbol{\xi}},t)
    -\boldsymbol{G}_r(\boldsymbol{\xi},t)\boldsymbol{\tau}_{\mathrm{F}}
    +\boldsymbol{d}_r(t),
    \label{relative_dynamic}
\end{equation}
where $\boldsymbol{f}_r$ accounts for the leader motion, follower dynamics, and moving-frame coupling, while $\boldsymbol{d}_r$ represents model uncertainties and environmental disturbances. Thus, these unknown terms need not be evaluated explicitly by the controller.

\subsection{Sliding Mode Controller}

Let $\boldsymbol{\xi}_d\in\mathbb{R}^6$ denote the prescribed relative pose in the follower frame and define
\begin{equation}
    \boldsymbol{e}=\boldsymbol{\xi}-\boldsymbol{\xi}_d.
\end{equation}
Because velocity measurements are unavailable, the relative velocity is reconstructed directly from the measured relative pose using the componentwise finite-time differentiator
\begin{align}
    \dot{\hat{\boldsymbol{\xi}}}
    &=
    \hat{\boldsymbol{v}}
    -\boldsymbol{\Lambda}_1
    |\tilde{\boldsymbol{\xi}}|^{1/2}
    \odot\operatorname{sgn}(\tilde{\boldsymbol{\xi}}), \\
    \dot{\hat{\boldsymbol{v}}}
    &=
    -\boldsymbol{\Lambda}_2
    \operatorname{sgn}(\tilde{\boldsymbol{\xi}}),
    \qquad
    \tilde{\boldsymbol{\xi}}
    =\hat{\boldsymbol{\xi}}-\boldsymbol{\xi},
    \label{eq:relative_differentiator}
\end{align}
where $\boldsymbol{\Lambda}_1$ and $\boldsymbol{\Lambda}_2$ are positive-definite diagonal matrices. For bounded relative acceleration and sufficiently large gains, $\hat{\boldsymbol{v}}$ converges to $\dot{\boldsymbol{\xi}}$ in finite time.

An integral-augmented sliding surface is defined as
\begin{equation}
    \boldsymbol{s}
    =
    \hat{\boldsymbol{v}}
    +\boldsymbol{C}_1\boldsymbol{e}
    +\boldsymbol{C}_2\int_0^t\boldsymbol{e}(\tau)\,\mathrm{d}\tau,
    \label{eq:sliding_surface}
\end{equation}
where $\boldsymbol{C}_1$ and $\boldsymbol{C}_2$ are positive-definite diagonal matrices. The output-feedback sliding mode control law is
\begin{equation}
    \boldsymbol{\tau}_{\mathrm{F}}
    =
    \widehat{\boldsymbol{G}}_r^{-1}
    \left(
        \boldsymbol{C}_1\hat{\boldsymbol{v}}
        +\boldsymbol{C}_2\boldsymbol{e}
        +\boldsymbol{\Gamma}\boldsymbol{s}
        +\boldsymbol{K}\operatorname{sgn}(\boldsymbol{s})
    \right),
    \label{eq:smc_control_law}
\end{equation}
where $\widehat{\boldsymbol{G}}_r$ is a nominal control-effectiveness matrix and $\boldsymbol{\Gamma}$ and $\boldsymbol{K}$ are positive-definite diagonal gain matrices. This control law depends only on the measured relative pose and does not require the follower velocity or explicit compensation for the hydrodynamic terms.

After the differentiator converges, the closed-loop sliding dynamics can be written as
\begin{equation}
    \dot{\boldsymbol{s}}
    =
    -\boldsymbol{\Gamma}\boldsymbol{s}
    -\boldsymbol{K}\operatorname{sgn}(\boldsymbol{s})
    +\boldsymbol{\Delta},
\end{equation}
where $\boldsymbol{\Delta}$ collects the bounded leader motion, model uncertainty, external disturbances, and input-gain mismatch. If
$|\Delta_i|\leq\bar{\Delta}_i$ and $k_i>\bar{\Delta}_i$, the Lyapunov function
$V=\frac{1}{2}\boldsymbol{s}^{\mathrm{T}}\boldsymbol{s}$ satisfies
\begin{equation}
    \dot{V}
    \leq
    -\boldsymbol{s}^{\mathrm{T}}\boldsymbol{\Gamma}\boldsymbol{s}
    -\sum_{i=1}^{6}
    \left(k_i-\bar{\Delta}_i\right)|s_i|
    \leq 0.
\end{equation}
Therefore, $\boldsymbol{s}$ reaches zero and the relative pose error converges to zero. In implementation, $\operatorname{sgn}(\cdot)$ is replaced by
$\operatorname{sat}(\cdot/\boldsymbol{\varphi})$ to reduce chattering, resulting in practical convergence within a tunable boundary layer.

\section{Experiments and Results}

We describe dataset construction, offline evaluation, and deployment on the follower OpenAUV. Closed-loop performance is evaluated under sliding mode control using the pool MoCap system.

\subsection{Dataset Construction}

\begin{figure}[thpb]
 \centering
 \subfigure[Moderate Light.]{
\label{moderate_illu}
\includegraphics[width=0.15\textwidth]{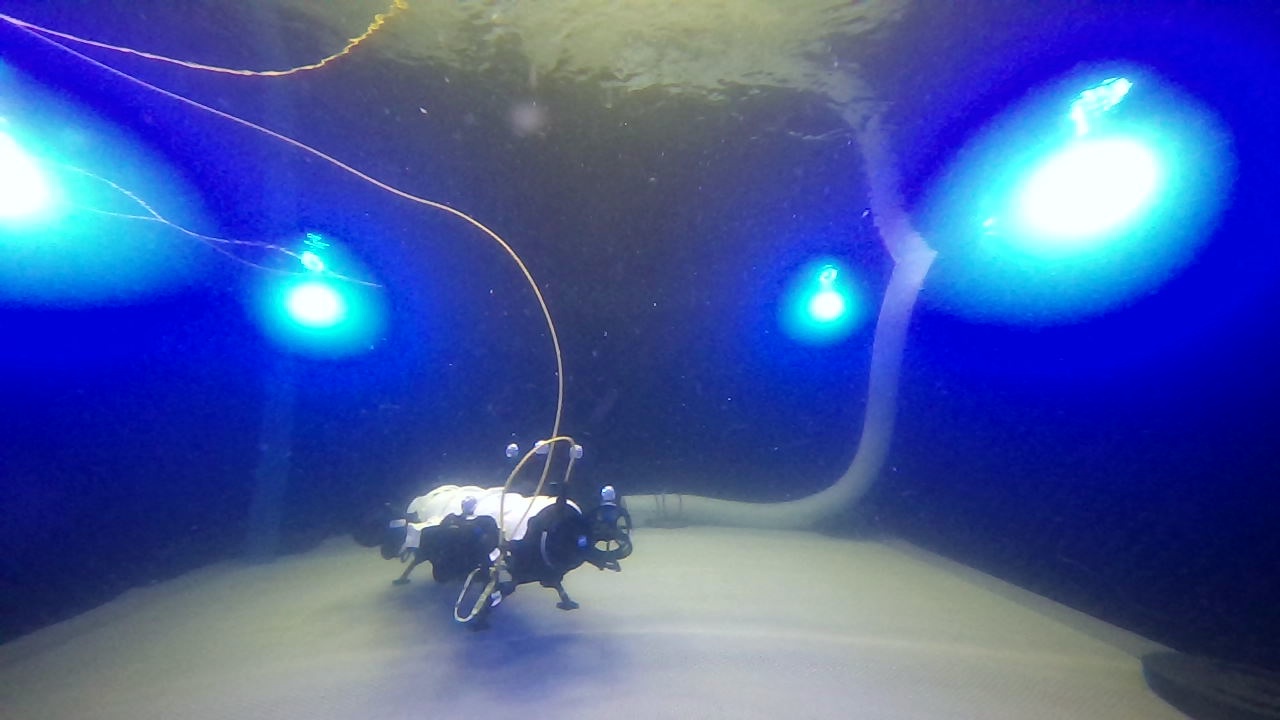}}
\subfigure[Dim Light.]{
\label{dim_illu}
\includegraphics[width=0.15\textwidth]{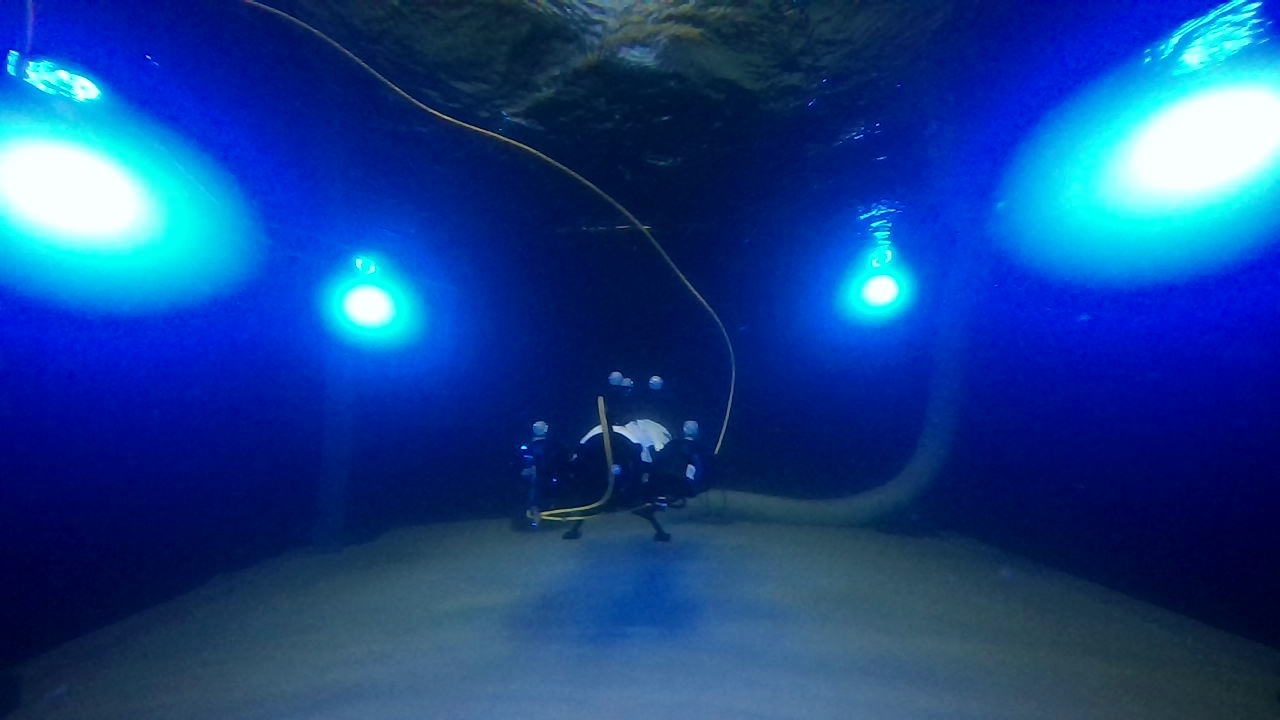}}
 \subfigure[Extreme Dim Light.]{
\label{edim_illu}
\includegraphics[width=0.15\textwidth]{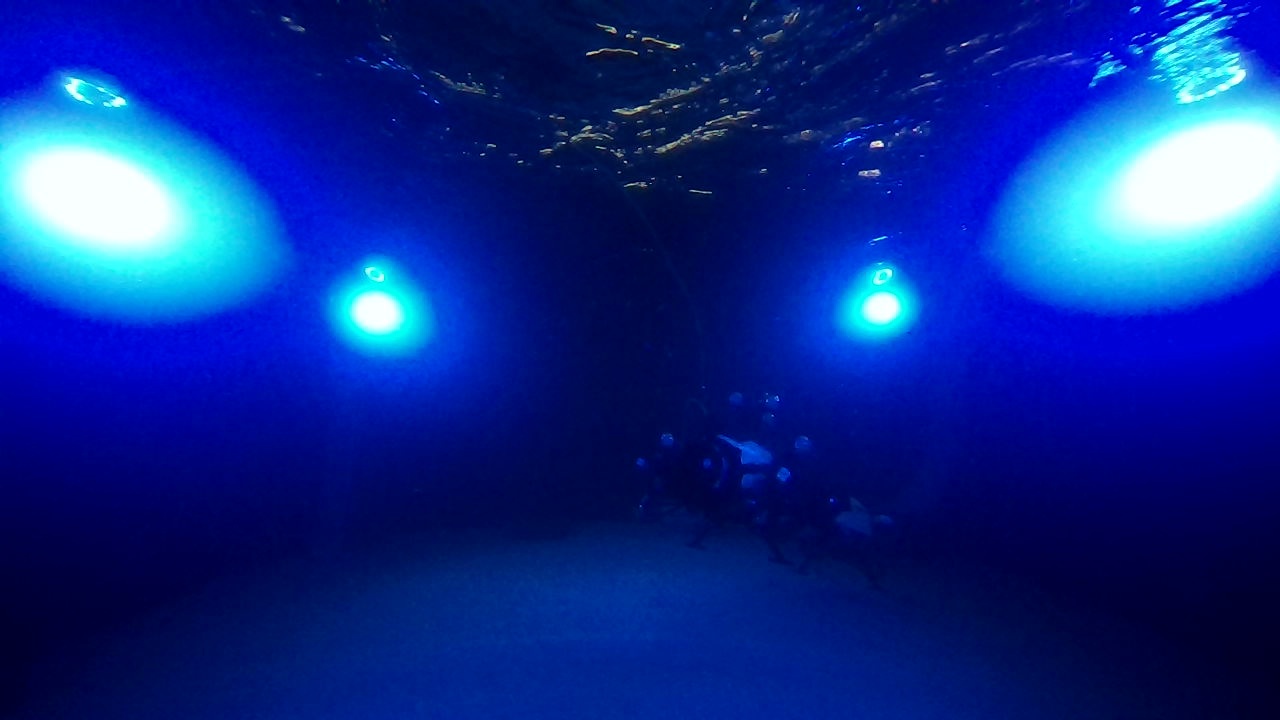}}
\caption{Illustrative images of illumination conditions. Fig.~\ref{moderate_illu}, Fig.~\ref{dim_illu}, and Fig.~\ref{edim_illu} illustrate the moderate, dim, and extremely dim light conditions during data collection, respectively. The illumination lights are located above the water surface; the blue lights visible in the images originate from the MoCap system cameras.}
\label{illu_cond}
\end{figure}

We collected 12 trials for each combination of two leader types, three illumination conditions, and nine trajectory patterns. During collection, a PID controller held the follower in hover while the leader followed predefined trajectories. The conditions are summarized in Table~\ref{data_collection} and Fig.~\ref{illu_cond}.

\begin{table}[htpb]
\caption{Dataset Collection Settings}
\vspace{-5pt}
\begin{center}
\renewcommand\arraystretch{1.4}
\begin{tabular}{cc}
\toprule
Variable & Description\\
\midrule
\multirow{2}*{Leader Vehicle Type} 
& 1. OpenAUV \\
& 2. BlueROV \\
\midrule
\multirow{3}*{Illumination Condition} 
& 1. Moderate Light \\
& 2. Dim Light \\ 
& 3. Extremely Dim Light \\
\midrule
\multirow{9}*{Leader Trajectory}  
& 1. Move Forward \\
& 2. Move Backward \\ 
& 3. Move Forward then Turn Left \\
& 4. Move Forward then Turn Right \\
& 5. Turn Left \\ 
& 6. Turn Right \\
& 7. Move Backward then Turn Left \\
& 8. Move Backward then Turn Right\\ 
& 9. Random Movement \\
\bottomrule
\end{tabular}
\begin{tablenotes}
\footnotesize
\item Twelve repeated trials were conducted  for each combination of conditions.
\end{tablenotes}
\end{center}
\label{data_collection}
\end{table}

The 648 trials contain all three sensing modalities. The camera sampled at 5 Hz and the acoustic and pressure sensors at 10 Hz. A multithreaded laptop process synchronized these streams with MoCap ground-truth poses.

\subsection{Offline Training and Testing}

Training has two stages. YOLO first localizes and segments the leader; IB-VE, PA-GATE, and KLD-WE then encode visual, acoustic, and pressure data for adaptive fusion and relative-pose regression.

\subsubsection{Stage One}

\begin{figure}[thpb]
 \centering
 \subfigure[Number of Instances per Class.]{
\label{num_instance}
\includegraphics[width=0.23\textwidth]{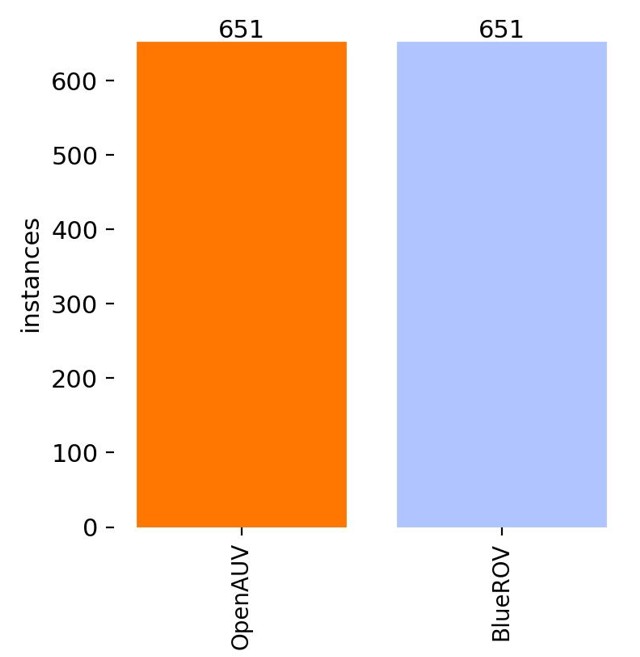}}
\subfigure[Overlapping Bounding Boxes with Aligned Centers.]{
\label{overlap_bbox}
\includegraphics[width=0.23\textwidth]{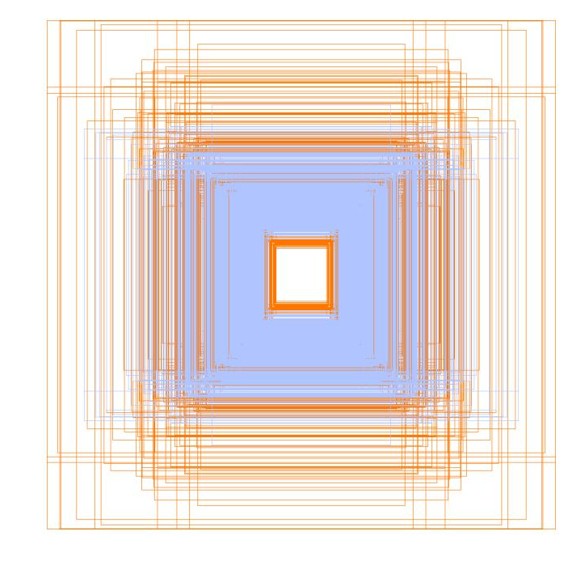}}
 \subfigure[Bounding Box Size Relative to Image.]{
\label{bbox_size}
\includegraphics[width=0.23\textwidth]{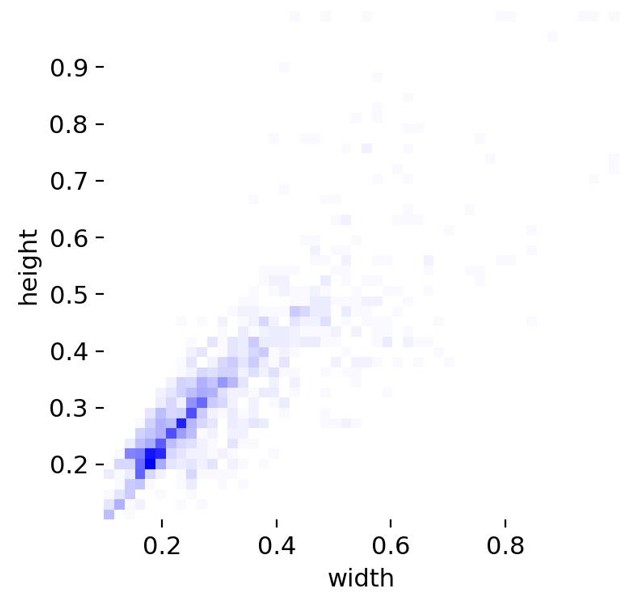}}
 \subfigure[Bounding Box Center Position Relative to Image.]{
\label{bbox_pos}
\includegraphics[width=0.23\textwidth]{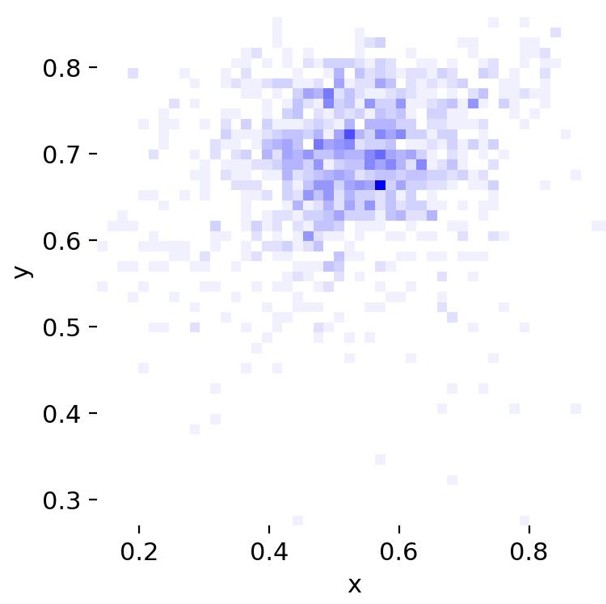}}
\caption{Training sample statistics. (a) Instance counts show a balanced class distribution. (b) Overlapping bounding boxes. (c) Bounding box size distribution relative to image dimensions. (d) Bounding box center position distribution relative to the image. The relative bounding box size distribution corresponds to the physical sizes of the OpenAUV and BlueROV, while the relative center positions indicate where the target typically appears within the camera's field of view.}
\label{AUVROV-label}
\end{figure}

We fine-tuned YOLOv11n-seg (2.83M parameters, 9.6 GFLOPs) on labeled images split into 651 training, 186 validation, and 93 test images per class. The class and box statistics are summarized in Figs.~\ref{AUVROV-label}, \ref{num_instance}, \ref{overlap_bbox}, \ref{bbox_size}, and \ref{bbox_pos}. The two classes are balanced; BlueROV boxes are smaller on average, consistent with its physical size. Normalized box dimensions are mostly 0.2--0.3, corresponding roughly to 1--2 m separation, and target locations cluster near the image center and lower half because vehicles were kept below the water surface.

With a batch size of 16 and training over 200 epochs, the model achieved the results in Table~\ref{YOLO-result}. During training and validation, four loss terms were considered: box loss, mask loss, class loss, and Distribution Focal Loss (DFL). The training procedure exhibited stable monotonic convergence without overfitting or instability. The training loss values converged to 0.7216 for bounding box loss, 1.3525 for segmentation mask loss, 1.0641 for classification loss, and 0.9653 for DFL, while the validation loss values were 0.6298, 1.1745, 1.0854, and 0.8908, respectively, indicating strong generalization ability.

The trained model achieved 0.9984 precision, 1.0000 recall, 0.9950 mAP@0.5, and 0.9488 mAP@0.5-0.95 for bounding box detection, and 0.9984 precision, 1.0000 recall, 0.9950 mAP@0.5, and 0.8879 mAP@0.5-0.95 for instance segmentation. The high consistency between box and mask metrics demonstrates the effectiveness and robustness of the trained model. Overall, the first-stage model yields high-precision segmentation with well-converged losses and robust performance that meets the requirements of the underwater application.

\begin{table}[htpb]
\caption{Stage One Model Evaluation}
\vspace{-5pt}
\begin{center}
\renewcommand\arraystretch{1.4}
\begin{tabular}{ccccc}
\toprule
\multirow{2}*{Train} & Box & Mask & Class & Distribution Focal \\
\cmidrule(lr){2-5} Loss & 0.7216 & 1.3525 & 1.0641 & 0.9653 \\
\midrule
\multirow{2}*{Validation} & Box & Mask & Class & Distribution Focal \\
\cmidrule(lr){2-5} Loss & 0.6298 & 1.1745 & 1.0854 & 0.8908 \\
\toprule
\multirow{2}*{Test} & Precision & Recall & mAP50 & mAP50-95 \\
\cmidrule(lr){2-5} (Box) & 0.9984 & 1.0000 & 0.9950 & 0.9488 \\
\midrule
\multirow{2}*{Test} & Precision & Recall & mAP50 & mAP50-95 \\
\cmidrule(lr){2-5} (Mask) & 0.9984 & 1.0000 & 0.9950 & 0.8879 \\
\bottomrule
\end{tabular}
\end{center}
\label{YOLO-result}
\end{table}

\begin{figure}[thpb]
 \centering
 \subfigure[Sample image of leader OpenAUV.]{
\label{sample_img}
\includegraphics[width=0.23\textwidth]{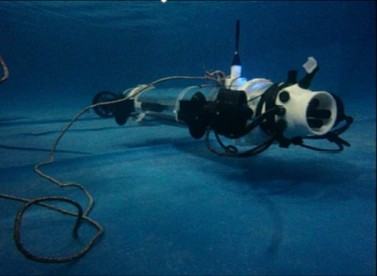}}
\subfigure[Visualization of constructed feature maps.]{
\label{f_maps}
\includegraphics[width=0.23\textwidth]{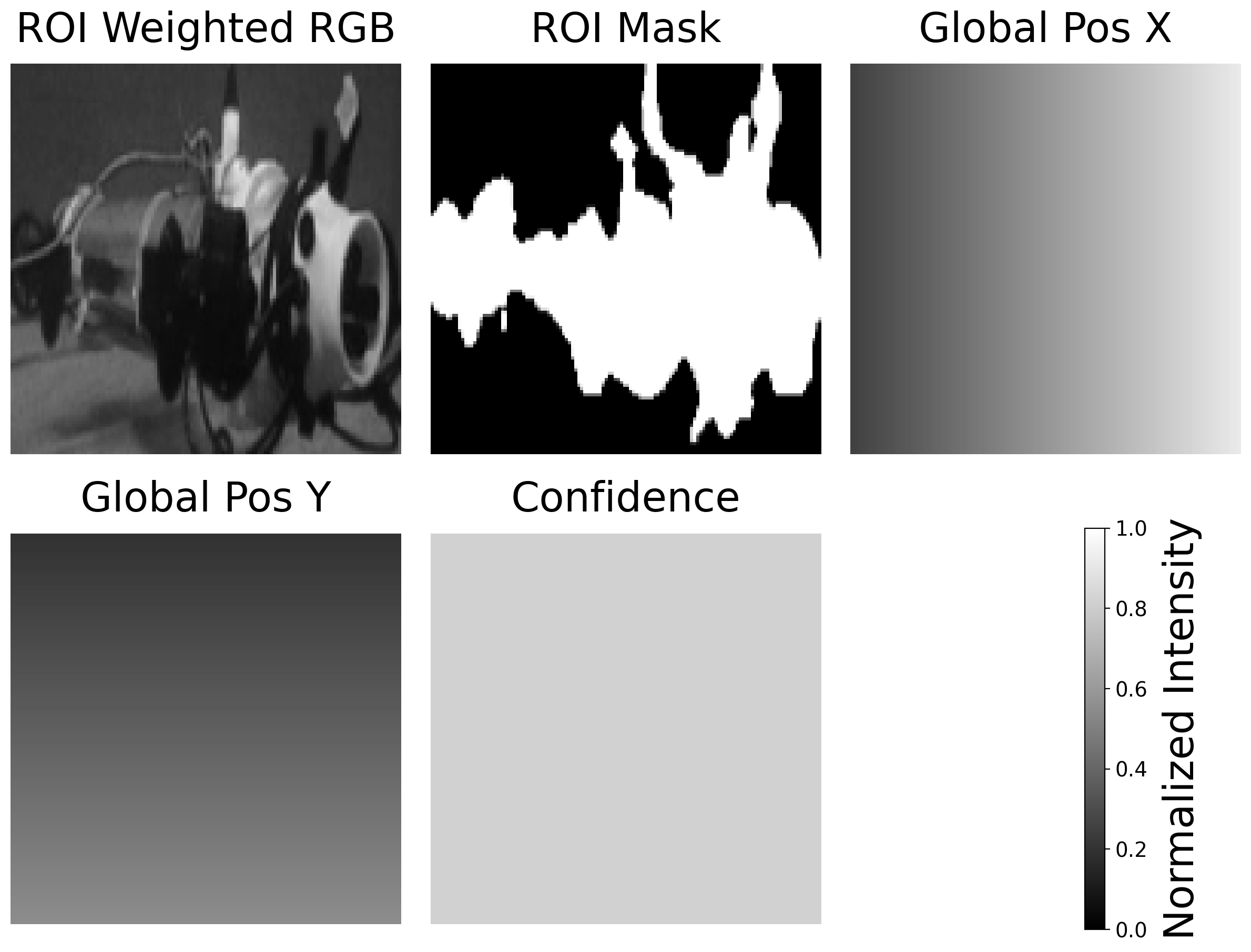}}
\caption{Constructed feature maps. (a) Sample image of the leader OpenAUV. (b) Five-channel feature map constructed from the first-stage results. The top row shows the weighted RGB, the rough segmentation mask, and the global position along the $x$-axis feature maps. The bottom row shows the global position along the $y$-axis, the confidence feature map, and the normalized intensity bar for the last three feature maps.}
\label{feature_map}
\end{figure}

Based on the detection and segmentation results, we constructed a five-channel feature map comprising a weighted RGB channel, a binary segmentation mask, global position encodings along the $x$- and $y$-axes, and detection confidence. The five channels contain, in order the weighted RGB, the rough target mask within the bounding box, the global image coordinates along $x$- and $y$-axes, and the prediction confidence. A sample image of the OpenAUV and the corresponding feature maps are visualized in Fig.~\ref{feature_map}. The top row shows the weighted RGB channel, the coarse segmentation mask, and the global $x$-coordinate map, while the bottom row shows the global $y$-coordinate map, the confidence map, and the normalized color scale for the last three feature maps.

\subsubsection{Stage Two}

In stage two, IB-VE, PA-GATE, KLD-WE, AWG, Mamba blocks, and the regression head were jointly trained end-to-end. PA-GATE used a model dimension of $d_\text{model}=32$. The batch size was 16 and training ran for 500 epochs. During network training, the number of Monte Carlo samples was set to $M=1$ per forward pass. The stochasticity inherent in mini-batch gradient descent across multiple epochs provides a sufficient approximation of the expectation, ensuring computational efficiency and low latency for deployment on the embedded edge device.

We evaluated the framework using mean absolute error (MAE), mean absolute percentage error (MAPE), and standard deviation (SD) for relative position $[\Delta x, \Delta y, \Delta z]$ and attitude $[\Delta \phi, \Delta \theta, \Delta \psi]$.

\begin{table*}[htpb]
\caption{Accuracy under Different Illumination (MAE and MAPE)}
\vspace{-5pt}
\begin{center}
\setlength{\tabcolsep}{3.7pt}
\renewcommand\arraystretch{1.4}
\begin{tabular}{ccccccccccccc}
\toprule
\multirow{2}*{Illumination} & \multicolumn{2}{c}{$\Delta x$} & \multicolumn{2}{c}{$\Delta y$} & \multicolumn{2}{c}{$\Delta z$} & \multicolumn{2}{c}{$\Delta \phi$} & \multicolumn{2}{c}{$\Delta \theta$} & \multicolumn{2}{c}{$\Delta \psi$} \\
\cmidrule(lr){2-3}\cmidrule(lr){4-5}\cmidrule(lr){6-7}\cmidrule(lr){8-9}\cmidrule(lr){10-11}\cmidrule(lr){12-13} & MAE(m) & MAPE(\%) & MAE(m) & MAPE(\%) & MAE(m) & MAPE(\%) & MAE($^\circ$) & MAPE(\%) & MAE($^\circ$) & MAPE(\%) & MAE($^\circ$) & MAPE(\%) \\
\midrule
Moderate & 0.057 & 3.8 & 0.038 & 3.8 & 0.013 & 3.3 & 0.537 & 5.4 & 0.779 & 3.9 & 5.119 & 4.3 \\
Dim & 0.060 & 4.0 & 0.042 & 4.2 & 0.011 & 2.8 & 0.503 & 5.0 & 0.837 & 4.2 & 5.786 & 4.8 \\
Extremely Dim & 0.067 & 4.5 & 0.044 & 4.4 & 0.014 & 3.5 & 0.652 & 6.5 & 1.153 & 5.8 & 6.338 & 5.3\\
\bottomrule
\end{tabular}
\end{center}
\label{framework-result}
\end{table*}

\begin{table}[htpb]
\caption{Robustness under Different Illumination (SD)}
\vspace{-5pt}
\begin{center}
\setlength{\tabcolsep}{5.2pt}
\renewcommand\arraystretch{1.4}
\begin{tabular}{ccccccc}
\toprule
Illumination & $\Delta x$(m) & $\Delta y$(m) & $\Delta z$(m) & $\Delta \phi$($^\circ$) & $\Delta \theta$($^\circ$) & $\Delta \psi$($^\circ$)\\
\midrule
Moderate & 0.044 & 0.032 & 0.006 & 0.414 & 0.607 & 4.738 \\
Dim & 0.051 & 0.025 & 0.008 & 0.457 & 0.637 & 3.082 \\
Extremely Dim & 0.101 & 0.033 & 0.015 & 0.582 & 1.071 & 5.615 \\
\bottomrule
\end{tabular}
\end{center}
\label{framework-SD}
\end{table}

\textbf{Overall Estimation Accuracy} Averaging the component-wise MAEs in Table~\ref{framework-result} across the three illumination conditions yields an average translational MAE of 3.8 cm and an average attitude MAE of 2.4$^\circ$. The corresponding component-wise MAEs, averaged across illumination conditions, were 6.1 cm longitudinally, 4.1 cm laterally, and 1.3 cm vertically. For attitude estimation, the corresponding MAEs were 0.564$^\circ$ for roll, 0.923$^\circ$ for pitch, and 5.748$^\circ$ for yaw. These results demonstrate centimeter-level relative-position estimation and identify yaw as the component with the largest absolute angular error.

\textbf{Robustness to Perceptual Degradation} To systematically evaluate the framework's robustness under visually degraded conditions, we analyzed performance across three illumination levels: moderate, dim, and extremely dim light. The quantitative results are summarized in Table~\ref{framework-result} (accuracy metrics) and Table~\ref{framework-SD} (robustness metrics).

Under dim light conditions, the framework achieved an average translational MAE of 3.77 cm, compared with 3.60 cm under moderate illumination. Under dim illumination, the lateral position error $\Delta y$ increased slightly from 3.8 cm to 4.2 cm, and the longitudinal error $\Delta x$ increased from 5.7 cm to 6.0 cm. The vertical error $\Delta z$ decreased slightly from 1.3 cm to 1.1 cm. For attitude estimation under dim light, the pitch error $\Delta \theta$ and yaw error $\Delta \psi$ increased from 0.779$^\circ$ to 0.837$^\circ$ and from 5.119$^\circ$ to 5.786$^\circ$, respectively, while roll error $\Delta \phi$ decreased slightly from 0.537$^\circ$ to 0.503$^\circ$. The standard deviation metrics in Table~\ref{framework-SD} reveal that the SD of the $\Delta y$ error decreased by 21.9\% and that of the $\Delta \psi$ error decreased by 35.0\% under dim conditions compared to moderate light. These counterintuitive improvements under dim lighting can be attributed to the adaptive weighting mechanism. As visual confidence decreases, the AWG dynamically redistributes fusion weights toward the acoustic and hydrodynamic modalities, which exhibit more consistent performance under reduced illumination. PA-GATE's physics-informed spatial encoding and KLD-WE's wake signature analysis become increasingly dominant information sources, effectively compensating for the degraded visual inputs.

Under extremely dim light conditions, the average translational MAE increased to 4.17 cm, representing a 15.7\% increase compared to moderate illumination. The longitudinal error $\Delta x$ increased by 17.5\%, the lateral error $\Delta y$ increased by 15.8\%, and the vertical error $\Delta z$ increased by 7.7\%. Attitude estimation exhibited more pronounced degradation: the roll error increased by 21.4\%, the pitch error increased substantially by 48.0\%, and the yaw error increased by 23.8\%. The mean orientation MAE increased from 2.145$^\circ$ to 2.714$^\circ$, corresponding to a 26.5\% overall degradation.

The robustness analysis in Table~\ref{framework-SD} further corroborates these findings. The standard deviation of the longitudinal position error increased by 129.5\% under extremely dim light, indicating significantly increased estimation variance. The vertical position SD increased by 150.0\%, and the pitch SD increased by 76.4\%. The yaw SD increased by 18.5\%. These increased SD values reflect the higher uncertainty inherent in relying predominantly on non-visual modalities when optical information becomes severely degraded. Nevertheless, the framework continues to provide relative-pose estimates even under extremely dim light, validating the effectiveness of the reliability-adaptive fusion strategy in dynamically shifting reliance toward acoustic and hydrodynamic modalities when visual confidence is low. This stands in stark contrast to purely vision-based approaches, which typically experience catastrophic failure under such conditions.

\textbf{Model Efficiency} As listed in Table~\ref{latency}, the proposed tri-modal fusion network requires 520.57M FLOPs and has 1.07M parameters. These metrics demonstrate that the framework is computationally lightweight and well suited for deployment on resource-constrained underwater vehicles. The linear-time complexity of the Mamba-SSM module, as opposed to the quadratic complexity of self-attention mechanisms, is instrumental in achieving this efficiency. Compared to a Transformer-based fusion framework requiring 1.67M parameters, the proposed framework reduces the parameter count by 35.9\% while maintaining superior estimation accuracy.

\subsubsection{Ablation Study}

To validate the individual contribution of each sensing branch, we conducted a modality-level ablation study by removing the acoustic branch (PA-GATE) and the hydrodynamic branch (KLD-WE) from the fusion pipeline in separate experiments. Together with the comparison study in Section~\ref{sec:comparison}, which isolates the adaptive weighting and temporal fusion components, these experiments cover the main architectural components of the framework. The quantitative results are presented in Table~\ref{ablation-result} (accuracy metrics) and Table~\ref{ablation-comparison-SD} (robustness metrics).

\begin{table*}[htpb]
\caption{Accuracy in the Ablation Study (MAE and MAPE)}
\vspace{-5pt}
\begin{center}
\setlength{\tabcolsep}{3.6pt}
\renewcommand\arraystretch{1.4}
\begin{tabular}{cccccccccccccc}
\toprule
\multirow{2}*{Structure} & \multicolumn{2}{c}{$\Delta x$} & \multicolumn{2}{c}{$\Delta y$} & \multicolumn{2}{c}{$\Delta z$} & \multicolumn{2}{c}{$\Delta \phi$} & \multicolumn{2}{c}{$\Delta \theta$} & \multicolumn{2}{c}{$\Delta \psi$} \\
\cmidrule(lr){2-3}\cmidrule(lr){4-5}\cmidrule(lr){6-7}\cmidrule(lr){8-9}\cmidrule(lr){10-11}\cmidrule(lr){12-13} & MAE(m) & MAPE(\%) & MAE(m) & MAPE(\%) & MAE(m) & MAPE(\%) & MAE($^\circ$) & MAPE(\%) & MAE($^\circ$) & MAPE(\%) & MAE($^\circ$) & MAPE(\%) \\
\midrule
w/o PA-GATE & 0.074 & 4.9 & 0.044 & 4.4 & 0.013 & 3.3 & 0.567 & 5.7 & 0.987 & 4.9 & 5.895 & 4.9\\
w/o KLD-WE & 0.075 & 5.0 & 0.047 & 4.7 & 0.015 & 3.8 & 0.608 & 6.1 & 1.013 & 5.5 & 6.638 & 5.5 \\
Complete & 0.068 & 4.5 & 0.040 & 4.0 & 0.012 & 3.0 & 0.531 & 5.3 & 0.887 & 4.4 & 5.613 & 4.7\\
\bottomrule
\end{tabular}
\end{center}
\label{ablation-result}
\end{table*}

\begin{table}[htpb]
\caption{Robustness in the Comparison and Ablation Studies (SD)}
\vspace{-5pt}
\begin{center}
\setlength{\tabcolsep}{5pt}
\renewcommand\arraystretch{1.4}
\begin{tabular}{ccccccc}
\toprule
Structure & $\Delta x$(m) & $\Delta y$(m) & $\Delta z$(m) & $\Delta \phi$($^\circ$) & $\Delta \theta$($^\circ$) & $\Delta \psi$($^\circ$)\\
\midrule
w/o PA-GATE & 0.111 & 0.066 & 0.018 & 0.769 & 1.441 & 9.877 \\
w/o KLD-WE & 0.109 & 0.071 & 0.022 & 0.840 & 1.477 & 10.194 \\
\midrule
Baseline 1 & 0.116 & 0.078 & 0.022 & 0.909 & 1.745 & 11.855 \\
Baseline 2 & 0.102 & 0.060 & 0.021 & 0.790 & 1.383 & 8.930 \\
\midrule
Ours & 0.097 & 0.057 & 0.018 & 0.743 & 1.340 & 9.108 \\
\bottomrule
\end{tabular}
\end{center}
\label{ablation-comparison-SD}
\end{table}

\textbf{Without PA-GATE} Removing PA-GATE and utilizing only visual information and pressure readings resulted in measurable degradation across all estimation dimensions. The longitudinal position MAE increased from 0.068 m to 0.074 m (8.8\% degradation), the lateral position MAE increased from 0.040 m to 0.044 m (10.0\% degradation), and the vertical position MAE increased from 0.012 m to 0.013 m (8.3\% degradation). Attitude estimation exhibited similar pattern of degradation. Roll MAE increased from 0.531$^\circ$ to 0.567$^\circ$ (6.8\% degradation), pitch MAE increased from 0.887$^\circ$ to 0.987$^\circ$ (11.3\% degradation), and yaw MAE increased from 5.613$^\circ$ to 5.895$^\circ$ (5.0\% degradation). These results demonstrate that PA-GATE is essential for interpreting sparse acoustic measurements. The physics-informed spatial encoding provides geometric constraints that resolve ambiguities in ranging measurements, particularly in the lateral direction where multipath effects are most pronounced.

The robustness metrics in Table~\ref{ablation-comparison-SD} further confirm the importance of PA-GATE. The standard deviation of longitudinal error increased from 0.097 m to 0.111 m (14.4\% increase), lateral error SD increased from 0.057 m to 0.066 m (15.8\% increase), pitch error SD increased from 1.340$^\circ$ to 1.441$^\circ$ (7.5\% increase), and yaw error SD increased from 9.108$^\circ$ to 9.877$^\circ$ (8.4\% increase). The consistent increase in SD across all dimensions indicates that PA-GATE not only improves mean accuracy but also enhances estimation stability by providing physically grounded regularization during temporal feature aggregation.

\textbf{Without KLD-WE} Excluding the KL-divergence-based wake encoder and removing hydrodynamic feature extraction from the fusion pipeline resulted in the most substantial performance degradation among all ablated configurations. The longitudinal position MAE increased from 0.068 m to 0.075 m (10.3\% degradation), the lateral position MAE increased from 0.040 m to 0.047 m (17.5\% degradation), and the vertical position MAE increased from 0.012 m to 0.015 m (25.0\% degradation). The degradation in attitude estimation was also pronounced. Roll MAE increased from 0.531$^\circ$ to 0.608$^\circ$ (14.5\%), pitch MAE increased from 0.887$^\circ$ to 1.013$^\circ$ (14.2\%), and yaw MAE increased from 5.613$^\circ$ to 6.638$^\circ$ (18.3\% degradation).

The robustness analysis reveals that removing KLD-WE  significantly weakens estimation stability. The longitudinal error SD increased from 0.097 m to 0.109 m (12.4\%), lateral error SD increased from 0.057 m to 0.071 m (24.6\%), vertical error SD increased from 0.018 m to 0.022 m (22.2\%), roll error SD increased from 0.743$^\circ$ to 0.840$^\circ$ (13.1\%), pitch error SD increased from 1.340$^\circ$ to 1.477$^\circ$ (10.2\%), and yaw error SD increased from 9.108$^\circ$ to 10.194$^\circ$ (11.9\%). These results conclusively demonstrate that hydrodynamic information extracted by KLD-WE plays a critical role in both accuracy and robustness. The KL-divergence constraint enables the encoder to learn distinctive wake signatures that are uniquely associated with the leader vehicle's relative pose, effectively complementing visual and acoustic information. Without this modality, the framework loses the ability to exploit flow-field patterns, particularly during leader maneuvers that generate characteristic hydrodynamic disturbances.

Taken together, the ablation results confirm that each module contributes distinct and complementary information to the fusion process. PA-GATE enhances acoustic interpretation through physics-informed encoding, while KLD-WE captures hydrodynamic signatures that are invariant to illumination changes. The complete framework, integrating all modules, achieves the optimal balance of accuracy and robustness.

\subsubsection{Comparison Study}
\label{sec:comparison}

We compared our adaptive fusion strategy against two common baselines used in multi-modal fusion.

\textbf{Baseline 1} A simple concatenation-based fusion approach, where the final feature vectors from each modality encoder are concatenated and fed directly into the regression head without adaptive weighting or temporal modeling.

\textbf{Baseline 2} A Transformer-based fusion architecture, where the Mamba-SSM blocks in our framework are replaced with standard multi-head self-attention Transformer blocks while retaining the adaptive weight generation mechanism.

Viewed as a component-level analysis, Baseline 1 isolates the joint contribution of the AWG and the temporal fusion module, whereas Baseline 2 isolates the contribution of the selective state space model by substituting it with Transformer blocks under an identical reliability-weighting front end. A finer-grained ablation that independently isolates IB-based compression in IB-VE, the KL regularization in KLD-WE, and the AWG from the temporal model requires retraining additional variants and is left to future work.

The quantitative comparison results for accuracy, robustness, and inference efficiency are presented in Table~\ref{comparison-result}, Table~\ref{ablation-comparison-SD}, and Table~\ref{latency}, respectively.

\begin{table*}[htpb]
\caption{Accuracy in the Comparison Study (MAE and MAPE)}
\vspace{-5pt}
\begin{center}
\setlength{\tabcolsep}{4.1pt}
\renewcommand\arraystretch{1.4}
\begin{tabular}{ccccccccccccc}
\toprule
\multirow{2}*{Structure} & \multicolumn{2}{c}{$\Delta x$} & \multicolumn{2}{c}{$\Delta y$} & \multicolumn{2}{c}{$\Delta z$} & \multicolumn{2}{c}{$\Delta \phi$} & \multicolumn{2}{c}{$\Delta \theta$} & \multicolumn{2}{c}{$\Delta \psi$} \\
\cmidrule(lr){2-3}\cmidrule(lr){4-5}\cmidrule(lr){6-7}\cmidrule(lr){8-9}\cmidrule(lr){10-11}\cmidrule(lr){12-13} & MAE(m) & MAPE(\%) & MAE(m) & MAPE(\%) & MAE(m) & MAPE(\%) & MAE($^\circ$) & MAPE(\%) & MAE($^\circ$) & MAPE(\%) & MAE($^\circ$) & MAPE(\%) \\
\midrule
Baseline 1 & 0.086 & 5.7 & 0.059 & 5.9 & 0.016 & 4.0 & 0.683 & 6.8 & 1.209 & 6.0 & 8.250 & 6.9 \\
Baseline 2 & 0.069 & 4.6 & 0.045 & 4.5 & 0.014 & 3.5 & 0.581 & 5.8 & 0.935 & 4.7 & 6.041 & 5.0 \\
Ours & 0.068 & 4.5 & 0.040 & 4.0 & 0.012 & 3.0 & 0.531 & 5.3 & 0.887 & 4.4 & 5.613 & 4.7\\
\bottomrule
\end{tabular}
\end{center}
\label{comparison-result}
\end{table*}

\textbf{Accuracy} The proposed framework consistently outperforms both baselines across all six estimation dimensions. Compared to Baseline 1, our approach achieves a 20.9\% reduction in longitudinal position MAE, a 32.2\% reduction in lateral position MAE, and a 25.0\% reduction in vertical position MAE. For attitude estimation, the improvements are equally substantial. Roll MAE improves by 22.3\%, pitch MAE improves by 26.6\%, and yaw MAE improves by 32.0\%. These results demonstrate that the simple concatenation baseline fails to effectively exploit the complementary nature of the three modalities, treating all features equally regardless of their information content or reliability under varying environmental conditions.

Compared to Baseline 2, the proposed Mamba-based framework achieves comparable or better accuracy with notable advantages in specific dimensions. The lateral position MAE improves by 11.1\%, vertical position MAE improves by 14.3\%, roll MAE improves by 8.6\%, pitch MAE improves by 5.1\%, and yaw MAE improves by 7.1\%. The longitudinal position MAE is comparable between the two approaches, with a 1.4\% difference. These results indicate that the selective state space modeling capabilities of Mamba-SSM are at least as effective as self-attention for temporal feature aggregation in this application, while offering significant computational advantages.

\textbf{Robustness} The robustness comparison in Table~\ref{ablation-comparison-SD} reveals that the proposed framework achieves the lowest average estimation variance across all configurations. Compared to Baseline 1, our approach reduces the longitudinal error SD by 16.4\%, lateral error SD by 26.9\%, roll error SD by 18.3\%, pitch error SD by 23.2\%, and yaw error SD by 23.2\%. These consistent improvements demonstrate that the adaptive weighting mechanism and the modality-specific feature encoding collectively enhance estimation stability, particularly under challenging operating conditions.

Compared to Baseline 2, the proposed framework reduces the longitudinal error SD by 4.9\%, lateral error SD by 5.0\%, roll error SD by 6.0\%, and pitch error SD by 3.1\%. The yaw error SD is comparable between the two approaches. These marginal but consistent improvements, combined with the substantial latency advantages, position the Mamba-based architecture as a compelling alternative to Transformer-based fusion for resource-constrained robotic applications.

\begin{table}[htpb]
\caption{Inference Latency and Model Information}
\vspace{-5pt}
\begin{center}
\renewcommand\arraystretch{1.4}
\begin{tabular}{cccc}
\toprule
Latency / Information & Baseline 1 & Baseline 2 & Ours \\
\midrule
Latency on RTX 4090 (ms) & 0.69 & 1.44  & 1.38 \\
Latency on Jetson AGX Orin (ms) & 6.54 & 47.48 & 17.62 \\
Params (M) & 0.45 & 1.67 & 1.07 \\
FLOPs (M) & 519.17 & 521.59 & 520.57 \\
\bottomrule
\end{tabular}
\begin{tablenotes}
\footnotesize
\item Inference on Jetson AGX Orin was optimized through TensorRT. RTX 4090 inference was not optimized.
\end{tablenotes}
\end{center}
\label{latency}
\end{table}

\textbf{Inference Latency} Table~\ref{latency} compares inference latency on desktop and embedded hardware. On NVIDIA RTX 4090 GPU, Baseline 1, Baseline 2, and the proposed framework achieved inference latencies of 0.69, 1.44, and 1.38 ms, respectively. All three approaches support real-time inference on this hardware.

\begin{figure}[!t]
    \centering
    \includegraphics[width=\columnwidth]{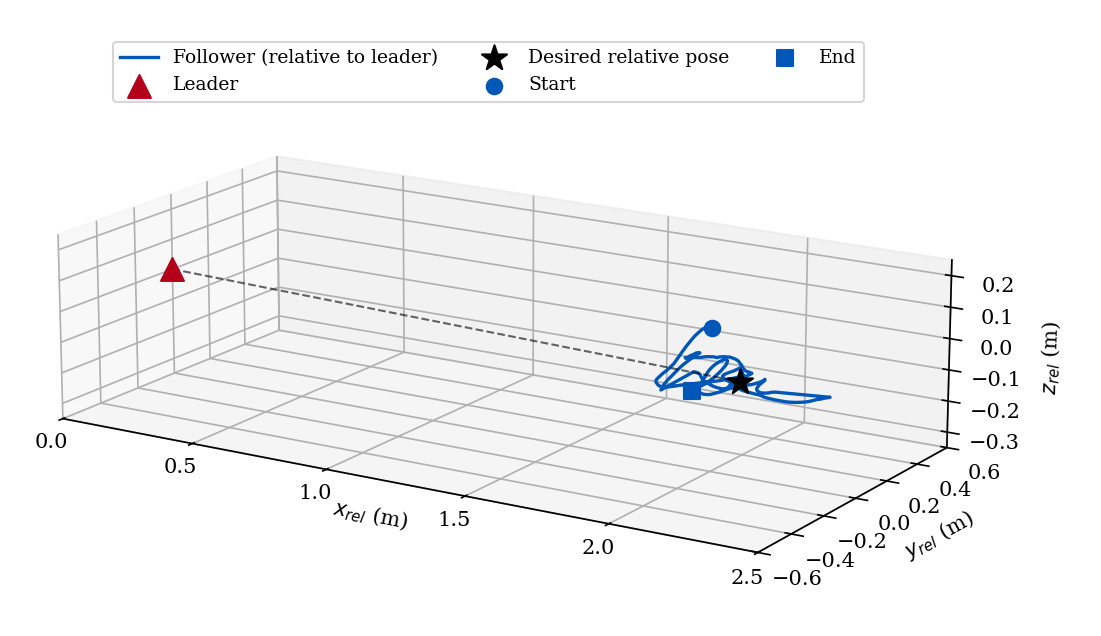}
    \caption{Follower trajectory in the leader-attached frame during the same segment. The leader is fixed at the origin; the star marks the desired relative pose $(2.10,0,0)\,\mathrm{m}$. The follower remains clustered around the desired pose throughout the maneuver.}
    \label{fig:closed_loop_3d}
\end{figure}

\begin{figure*}[!t]
    \centering
    \includegraphics[width=0.95\textwidth]{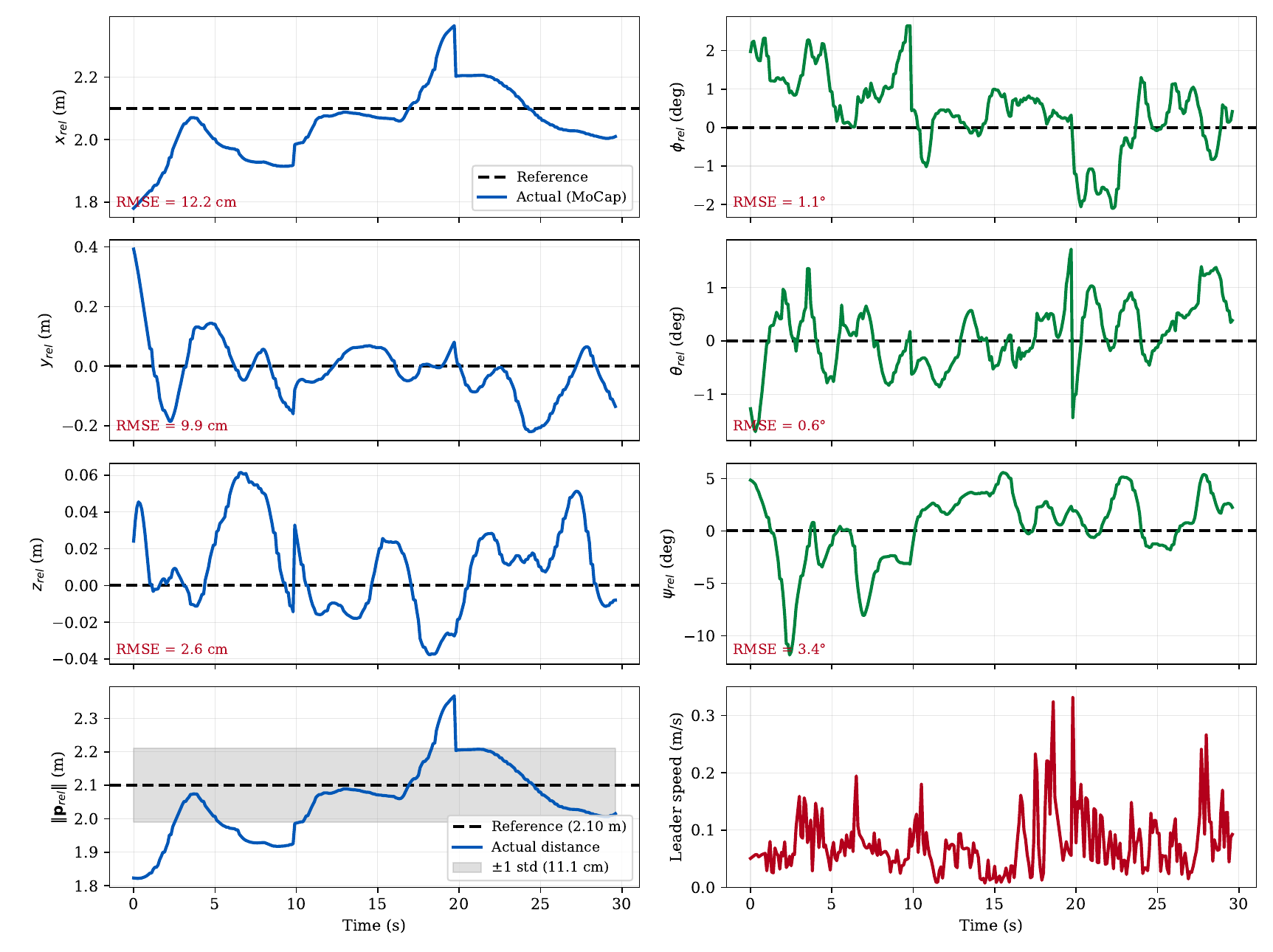}
    \caption{Closed-loop relative-pose and attitude tracking over a representative $30\,\mathrm{s}$ segment with the leader in motion. Left column: relative position $x_{\mathrm{rel}}$, $y_{\mathrm{rel}}$, $z_{\mathrm{rel}}$ and relative distance $\|\mathbf{p}_{\mathrm{rel}}\|$ (solid) against their references (dashed). Right column: relative roll, pitch, yaw, and the leader speed. All measured quantities were obtained from the motion capture system.}
    \label{fig:closed_loop_tracking}
\end{figure*}

However, the performance gap widens dramatically on the embedded NVIDIA Jetson AGX Orin edge device. Baseline 1 achieves 6.54 ms latency due to its minimal architecture, Baseline 2 has a latency of 47.48 ms due to the quadratic complexity of self-attention, and the proposed framework achieves 17.62 ms latency. This represents a 2.69-fold speedup over the Transformer-based Baseline 2, while maintaining superior estimation accuracy. The proposed sensing framework achieves real-time throughput on the Jetson AGX Orin and enables real-time closed-loop control. The proposed framework also uses 35.9\% fewer parameters than the Transformer-based baseline.

\subsection{Edge Deployment and Online Validation}

\subsubsection{Experiments in a Controlled Laboratory Pool}

In this section, we describe the deployment of the multi-modal perception algorithm on the OpenAUV platform and validate the leader-follower system through experiments in a pool.

\subsubsection{Deployment of Stage 1 of the Perception Algorithm}

The PyTorch checkpoint was exported to a fixed-input ONNX graph and compiled as an FP16 TensorRT engine. Layer fusion and Orin-specific CUDA kernel optimization reduce runtime overhead. This segmentation pipeline reduced latency from 45.20 ms to 27.37 ms (1.65-fold speedup, 36.54 FPS).

\subsubsection{Deployment of Stage 2 of the Perception Algorithm}

The second-stage fusion model was exported to fixed-shape ONNX and compiled as an FP16 TensorRT engine. Direct binding of CUDA tensor memory to the execution context avoids extra I/O copies. Latency decreased from approximately 300 ms to 17.62 ms (17.03-fold speedup, 56.75 FPS), with negligible change in positional accuracy and comparable angular precision.

We validated the leader-follower system in the $4\,\mathrm{m}\times2\,\mathrm{m}$ laboratory pool to quantitatively evaluate the closed-loop performance of the proposed framework.

\textbf{Hardware Setup and Loop Frequency.} The perception framework was deployed on an NVIDIA Jetson AGX Orin integrated with the sliding mode controller (SMC), forming a closed-loop perception and control system on the follower OpenAUV. The fused multimodal relative-pose estimate was updated at $5\,\mathrm{Hz}$, while the SMC executed at a fixed period of $100\,\mathrm{ms}$ ($10\,\mathrm{Hz}$).

\textbf{Trajectory Tracking Accuracy and Robustness.} With the leader OpenAUV following a predefined trajectory, the follower was commanded to hold a constant relative pose of $(2.10,\,0,\,0)\,\mathrm{m}$ with zero relative attitude, while the underwater motion capture system recorded the ground truth. A representative $30\,\mathrm{s}$ segment of the follower trajectory in the leader-attached frame is visualized in Fig.~\ref{fig:closed_loop_3d}, with the corresponding position and attitude curves shown in Fig.~\ref{fig:closed_loop_tracking}. The follower maintained a standard deviation of $0.11\,\mathrm{m}$ around the $2.10\,\mathrm{m}$ reference. The  position RMSE values were $12.2\,\mathrm{cm}$, $9.9\,\mathrm{cm}$, and $2.6\,\mathrm{cm}$ in the longitudinal, lateral, and vertical directions, respectively, and the relative-attitude RMSE did not exceed $1.1^\circ$, $0.6^\circ$, and $3.4^\circ$ for roll, pitch, and yaw. The follower thus preserved the formation geometry under continuous leader motion, confirming that the lightweight perception front end provides state estimates of sufficient quality for closed-loop formation control on resource-constrained edge hardware.

\subsection{Discussion}

Performance did not vary monotonically with illumination. Extremely dim light degraded estimates, whereas dim light improved several metrics relative to moderate light. Reliability-adaptive weighting may reduce the visual contribution as image quality falls, although this trend is specific to the tested pool conditions.

The hydrodynamic encoder is less affected by illumination because wake-induced pressure fluctuations do not depend on visibility. Acoustic ranges add geometric constraints complementary to vision and pressure sensing. Open-water tests remain necessary to evaluate currents, turbidity changes, and larger separation distances.

\section{Conclusion}

This paper presents reliability-adaptive fusion of visual, acoustic, and hydrodynamic sensing for continuous 6-DoF relative-pose tracking between moving underwater robots. IB-VE, PA-GATE, and KLD-WE encode the three modalities; a reliability estimator weights their features before Mamba-SSM fusion. In pool experiments, the framework achieves a mean translational error of 3.8 cm and mean orientation error of 2.4$^\circ$. With 1.07M parameters and 520.57M FLOPs, the fusion network runs in 17.62 ms on Jetson AGX Orin, a 2.69-fold speedup over the Transformer baseline. Closed-loop tests demonstrate formation maintenance using onboard perception.

Future work will extend closed-loop validation to larger environments and a wider range of leader motions and sensing conditions. We will also explore online learning and domain adaptation to accommodate previously unseen leader types and changing environmental conditions without requiring complete offline retraining.

\bibliography{ref}

\vfill

\end{document}